\documentclass[10pt,journal,compsoc]{IEEEtran}
\ifCLASSOPTIONcompsoc
  \usepackage[nocompress]{cite}
\else
  \usepackage{cite}
\fi

\usepackage[numbers]{natbib}
\usepackage[pdftex]{graphicx}
\usepackage{amsmath, dsfont}
\usepackage{mathtools}
\DeclarePairedDelimiterX{\norm}[1]{\lVert}{\rVert}{#1}
\usepackage{balance}
\usepackage{xcolor}

\newcommand{\revision}[1]{\textcolor{black}{#1}}
\newcommand{\newrevision}[1]{\textcolor{black}{#1}}
\usepackage{array}

\ifCLASSOPTIONcompsoc
 \usepackage[caption=false,font=footnotesize,labelfont=sf,textfont=sf]{subfig}
\else
 \usepackage[caption=false,font=footnotesize]{subfig}
\fi

\usepackage{url}
\usepackage{amssymb}
\usepackage{bbm}
\usepackage{bm}
\usepackage{multirow}
\usepackage{booktabs}
\begin{document}
%
% paper title
% Titles are generally capitalized except for words such as a, an, and, as,
% at, but, by, for, in, nor, of, on, or, the, to and up, which are usually
% not capitalized unless they are the first or last word of the title.
% Linebreaks \\ can be used within to get better formatting as desired.
% Do not put math or special symbols in the title.
\title{Self-Supervised Learning of Graph Neural Networks: A Unified Review}

\author{Yaochen Xie,
        Zhao Xu,
        Jingtun Zhang,
        Zhengyang Wang,
        Shuiwang Ji,~\IEEEmembership{Senior Member,~IEEE}
\IEEEcompsocitemizethanks{\IEEEcompsocthanksitem Y. Xie, Z. Xu, J. Zhang, and S. Ji are with Department of Computer Science \& Engineering,
Texas A\&M University, College Station,
TX 77843.\protect\\
% note need leading \protect in front of \\ to get a newline within \thanks as
% \\ is fragile and will error, could use \hfil\break instead.
E-mail: \{ethanycx, zhaoxu, zjt6791, sji\}@tamu.edu
\IEEEcompsocthanksitem Z. Wang is with Amazon.com Services LLC, Seattle, WA 98109.\protect\\
E-mail: zhengywa@amazon.com}% <-this % stops an unwanted space
%\thanks{Manuscript received xxx; revised xxx}
}

% The paper headers
\markboth{}{}

% for Computer Society papers, we must declare the abstract and index terms
% PRIOR to the title within the \IEEEtitleabstractindextext IEEEtran
% command as these need to go into the title area created by \maketitle.
% As a general rule, do not put math, special symbols or citations
% in the abstract or keywords.
\IEEEtitleabstractindextext{
\begin{abstract}
Deep models trained in supervised mode have achieved remarkable
success on a variety of tasks. When labeled samples are limited,
self-supervised learning (SSL) is emerging as a new paradigm for
making use of large amounts of unlabeled samples. SSL has achieved
promising performance on natural language and image learning tasks.
Recently, there is a trend to extend such success to graph data
using graph neural networks (GNNs). In this survey, we provide a
unified review of different ways of training GNNs using SSL. Specifically, we
categorize SSL methods into contrastive and predictive models. In either category, we
provide a unified framework for methods as well as how these methods
differ in each component under the framework. Our unified treatment of
SSL methods for GNNs sheds light on the similarities and differences
of various methods, setting the stage for developing new methods and
algorithms. We also summarize different SSL settings and the
corresponding datasets used in each setting. To facilitate
methodological development and empirical comparison, we develop a standardized testbed for SSL in GNNs, including
implementations of common baseline methods, datasets, and evaluation
metrics.
\end{abstract}

% Note that keywords are not normally used for peerreview papers.
}

% make the title area
\maketitle

% To allow for easy dual compilation without having to reenter the
% abstract/keywords data, the \IEEEtitleabstractindextext text will
% not be used in maketitle, but will appear (i.e., to be "transported")
% here as \IEEEdisplaynontitleabstractindextext when the compsoc
% or transmag modes are not selected <OR> if conference mode is selected
% - because all conference papers position the abstract like regular
% papers do.
\IEEEdisplaynontitleabstractindextext
% \IEEEdisplaynontitleabstractindextext has no effect when using
% compsoc or transmag under a non-conference mode.

% For peer review papers, you can put extra information on the cover
% page as needed:
% \ifCLASSOPTIONpeerreview
% \begin{center} \bfseries EDICS Category: 3-BBND \end{center}
% \fi
%
% For peerreview papers, this IEEEtran command inserts a page break and
% creates the second title. It will be ignored for other modes.
\IEEEpeerreviewmaketitle

\IEEEraisesectionheading{\section{Introduction}\label{sec:introduction}}

% 1. The neural networks, the supervised learning
\IEEEPARstart{A} deep model takes some data as its inputs and is trained to output desired predictions. A common way to train a deep model is to use the supervised mode in which a sufficient amount of input data and label pairs are given. However, since a large number of labels are required, the supervised training becomes inapplicable in many real-world scenarios, where labels are expensive, limited, \revision{imbalanced~\cite{yang2020rethinking}, }or even unavailable.
% 2. With a huge amount of unlabeled data, self-supervised learning
In such cases, self-supervised learning (SSL) enables the training of deep models on unlabeled data, removing the need of excessive annotated labels. When no labeled data is available, SSL serves as an approach to learn representations from unlabeled data itself. When a limited number of labeled data is available, SSL from unlabeled data can be used either as a pre-training process after which labeled data are used to fine-tune the pre-trained deep models for downstream tasks, or as an auxiliary training task that contributes to the performance of main tasks.

% 3. SSL studies in general
Recently, SSL has shown its promising capability in data restoration tasks, such as image super-resolution~\cite{Ulyanov2018deep}, image denoising~\cite{xie2020noise2same, laine2019high, krull2019noise2void}, and single-cell analysis~\cite{batson2019noise2self}. It has also achieved remarkable progress in representation learning for different data types, including language sequences~\cite{devlin2019bert, wu2019self, wang2019self}, images~\cite{hassani2020contrastive, chen2020simple, oord2018representation,tian2019contrastive}, and graphs with sequence models~\cite{Perozzi2014deepwalk, narayanan2017graph2vec} or spectral models~\cite{tsitsulin2018sgr}. The key idea of these methods is to define pretext training tasks to capture and use the dependencies among different dimensions of the input data, \emph{e.g.}, the spatial, temporal, or channel dimensions, with robustness and smoothness. Taking the image domain as an example, \citet{doersch2015unsupervised, noroozi2016unsupervised}, and \citet{he2020momentum} design different pretext tasks to train convolutional neural networks (CNNs) to capture relationships between different crops from an image. \citet{chen2020simple} and \citet{grill2020bootstrap} train CNNs to capture dependencies between different augmentations of an image.

\begin{figure}
    \centering
    \includegraphics[width=0.48\textwidth]{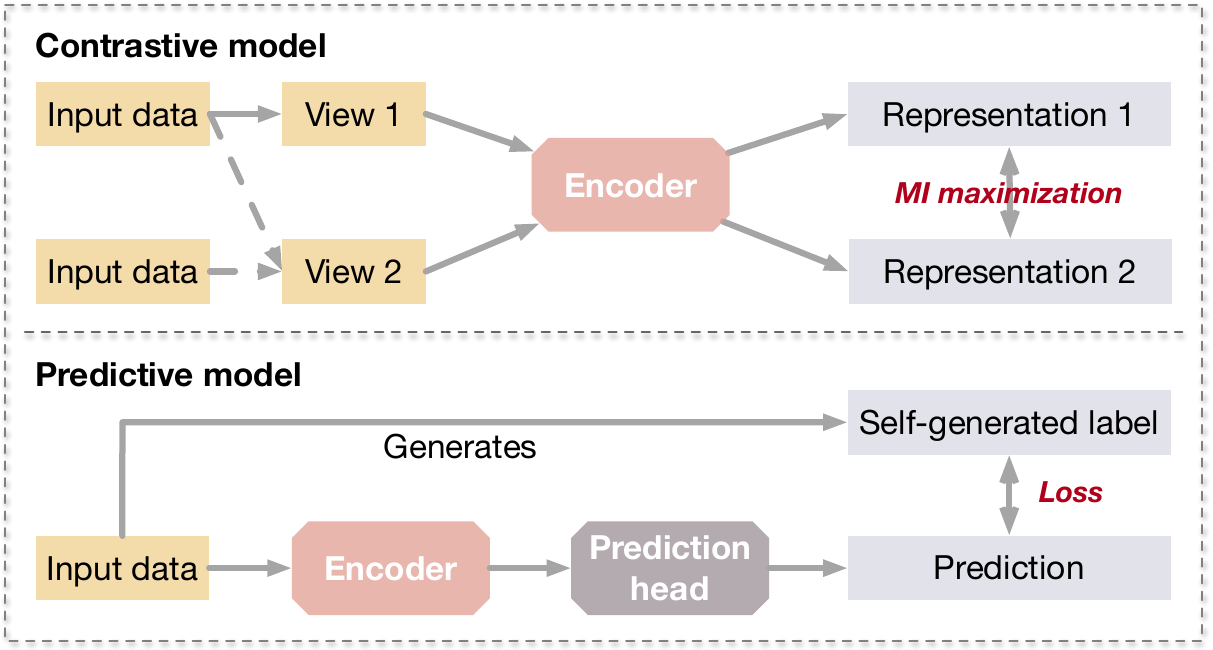}
    \caption{A comparison between the contrastive model and the predictive model in general.}
    \label{fig:con_vs_pre}
\end{figure}

% 4. two types: predictive and contrastive
Based on how the pretext training tasks are designed, SSL methods can be divided into two categories; namely contrastive models and predictive models. The major difference between the two categories is that contrastive models require data-data pairs for training, while predictive models require data-label pairs, where the labels are self-generated from the data, as illustrated in Figure~\ref{fig:con_vs_pre}. Contrastive models usually utilize self-supervision to learn data representation or perform pre-training for downstream tasks. Given the data-data pairs, contrastive models perform discrimination between positive pairs and negative pairs. On the other hand, predictive models are trained in a supervised fashion, where the labels are generated based on certain properties of the input data or by selecting certain parts of the data. Predictive models usually consist of an encoder and one or more prediction heads. When applied as a representation learning or pre-training method, the prediction heads of a predictive model are removed in the downstream task.

% 5. gnns
In graph data analysis, SSL can potentially be of great importance to make use of a massive amount of unlabeled graphs such as molecular graphs~\cite{wu2018moleculenet, wang2020moleculekit}. With the rapid development of graph neural networks (GNNs)~\cite{kipf2017semi,xu2018how,Gao:ICML19,liu2020non, cai2020line, Liu:Deeper,CaiMlinkAAAI20}, basic components of GNNs~\cite{Gao:KDD18,Gao:www19,gao2020topology,wang2020second,Yuan:ICLR2020,gao2019graph} and other related fields~\cite{Yuan:XGNN,Liu:Protein} have been well studied and made substantial progress. In comparison, applying SSL on GNNs is still an emerging field.
% 5. For graphs, the core challenges
Due to the similarity in data structure, many SSL methods for GNNs are inspired by methods in the image domain, such as DGI~\cite{velikovi2019deep} and graph autoencoders~\cite{kipf2016variational}. However, there are several key challenges in applying SSL on GNNs due to the uniqueness of the graph-structured data. To obtain good representations of graphs and perform effective pre-training, self-supervised models are supposed to capture essential information from both nodes attributes and structural topology of graphs~\cite{ma2020deep}. For contrastive models, as the GPU memory issue of performing self-supervised learning is not a major concern for graphs, the key challenge lies in how to obtain good views of graphs and the selection of graph encoder for different models and datasets. For predictive models, it becomes essential that what labels should be generated so that the non-trivial representations are learned to capture information in both node attributes and graph structures.

\begin{figure*}[t]
    \centering
    \includegraphics[width=0.98\textwidth]{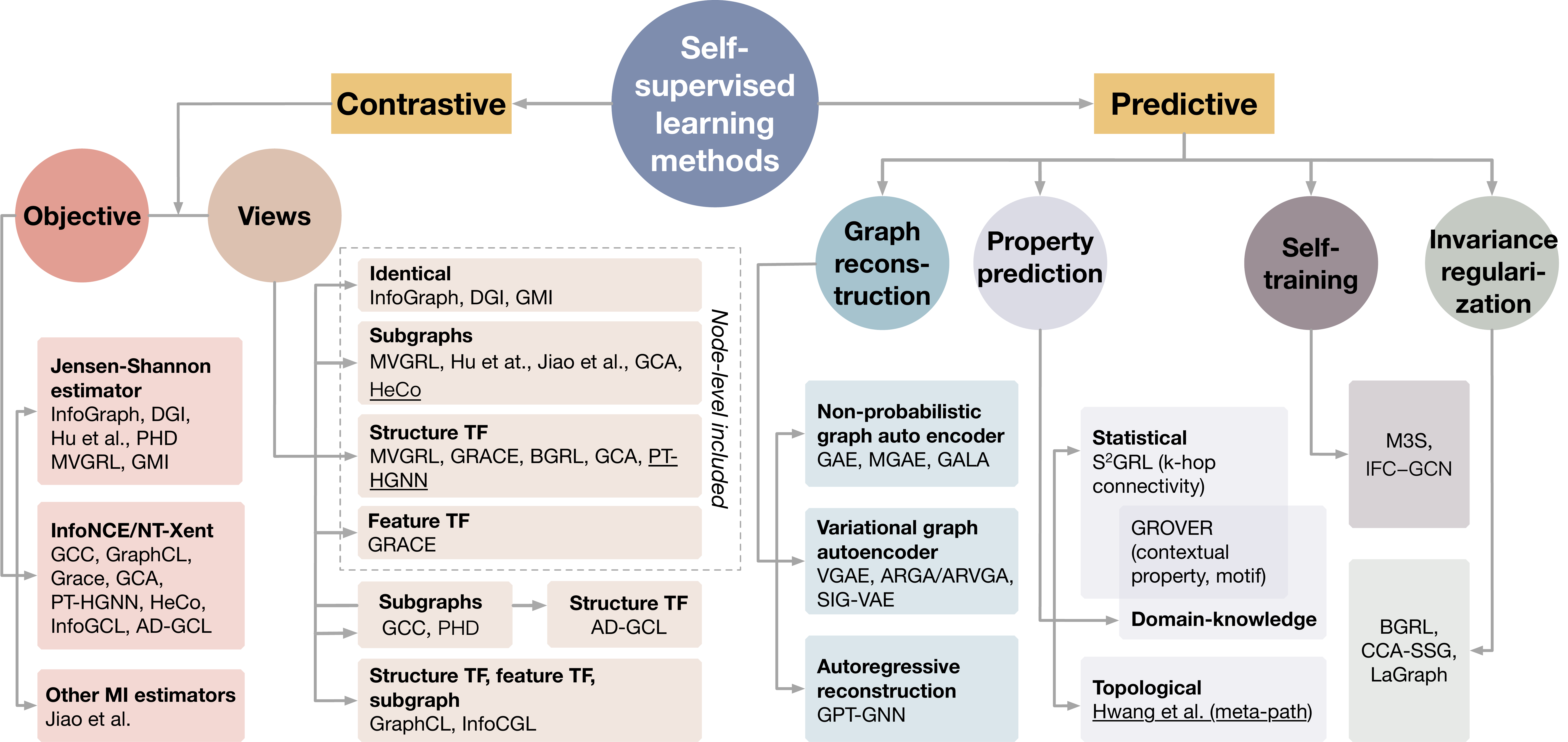}
    \caption{An overview of self-supervised learning methods. We categorize self-supervised learning methods into two branches: contrastive methods and predictive methods. For contrastive methods, we further divide them regarding either views generation or objective. From the aspect of views generation, Infograph~\cite{sun2019infograph}, DGI~\cite{velikovi2019deep} and GMI~\cite{peng2020graph} contrast views between nodes and graph; Hu et al.~\cite{Hu2020Strategies} and Jiao et al.~\cite{jiao2020sub} contrast views between nodes and subgraph; MVGRL~\cite{hassani2020contrastive} and GCA~\cite{zhu2021graph} contrast views between nodes and subgraph or structurally transformed graph; GRACE~\cite{zhu2020deep} and BGRL~\cite{thakoor2021bootstrapped} contrast views between nodes and structurally transformed graph or featurally transformed graph. Above methods include node-level representation to generate local/global contrastive pairs. Dissimilarly, following methods use global representation only to generate global/global contrastive pairs. GCC~\cite{qiu2020gcc} contrasts views between subgraphs; GraphCL~\cite{you2020graph} contrasts views of subgraphs and randomly transformed graphs. From aspect of objective, Infograph~\cite{sun2019infograph}, DGI~\cite{velikovi2019deep}, Hu et al.~\cite{Hu2020Strategies}, MVGRL~\cite{hassani2020contrastive} and GMI~\cite{peng2020graph} employ Jensen-Shannon estimator; GCC~\cite{qiu2020gcc}, GraphCL~\cite{you2020graph}, GRACE~\cite{zhu2020deep} and GCA~\cite{zhu2021graph} employ InfoNCE (NT-Xent); Jiao et al.~\cite{jiao2020sub} use other MI estimators. For the predictive methods, we further divide them into graph reconstruction, property prediction, self-training, and invariance regularization methods. Under graph reconstruction, GAE~\cite{kipf2016variational}, MGAE~\cite{wang2017mgae}, and GALA~\cite{park2019symmetric} utilize the non-probabilistic graph autoencoder; VGAE~\cite{kipf2016variational}, ARGA/ARVGA~\cite{pan2018adversarially}, and SIG-VAE~\cite{hasanzadeh2019semi} utilize variational graph autoencoder; GPT-GNN~\cite{hu2020gpt} applies autoregressive reconstruction. Under property prediction, S$^2$GRL~\cite{peng2020self} performs the prediction of k-hop connectivity as a statistical property; GROVER~\cite{rong2020grover} performs predictions of a statistical contextual property and a domain-knowledge involved property; \citet{hwang2020self} predict a topological property, meta-path. M3S~\cite{sun2020multi} and ICF-GCN~\cite{hu2021rectifying} employs self-training and node clustering to provide self-supervision.  BGRL~\cite{thakoor2021bootstrapped}, CCA-SSG~\cite{zhang2021canonical}, and LaGraph~\cite{xie2022self} derive self-supervised objectives involving invariance regularization without requiring negative pairs. \newrevision{SSL methods for heterogeneous graphs are marked with underlines. We discuss and summarize SSL methods for heterogeneous graphs and dynamic graphs in Appendix~A. We further discuss and compare contrastive and predictive methods in Appendix~B.}}
    \label{fig:overview}
\end{figure*}

To foster methodological development and facilitate empirical comparison, we review SSL methods of GNNs and provide unified views for both contrastive and predictive methods. Our unified treatment of this topic may shed light on the similarities and differences among current methods and inspire new methods. We also provide a standardized testbed as a convenient and flexible open-source platform for performing empirical comparisons. We summarize the contributions of this survey as follows:
\begin{itemize}
    \item We provide thorough and up-to-date reviews on SSL methods for graph neural networks. To the best of our knowledge, our survey presents the first review of SSL specifically on graph data.
    \item We unify existing contrastive learning methods for GNNs with a general framework. Specifically, we unify the contrastive objectives from the perspective of mutual information. From this fresh view, different ways to perform contrastive learning can be considered as performing three transformations to obtain views. We review theoretical and empirical studies and provide insights to guide the choice of each component in the framework.
    \item We categorize and unify SSL methods with self-generated labels as predictive learning methods, and elucidate their connections and differences by different ways of obtaining labels.
    \item We summarize common settings of SSL tasks and commonly used datasets of various categories under different settings, setting the stage for developments of future methods.
    \item We develop a standardized testbed for applying SSL on GNNs, including implementations of common baseline methods and benchmarks, enabling convenient and flexible customization for future methods.
\end{itemize}
An overview of self-supervised learning methods of different categories is given in Figure~\ref{fig:overview}.

\revision{A recent work~\cite{liu2020self} provides thorough and general literature reviews on self-supervised learning for vision, natural language processing, and graph mining tasks. While both~\cite{liu2020self} and our work review SSL methods as contrastive ones and non-contrastive ones, we distinguish our reviews from~\cite{liu2020self} by the following distinct differences.}
\begin{itemize}
    \item \revision{\citet{liu2020self} and our paper propose different taxonomies for SSL methods from different aspects of view. Specifically, \cite{liu2020self} categorizes contrastive methods by the levels of contrast such as instance-instance and context-instance, where mutual-information is considered as one specific subcategory at the context-instance level. In contrast, our taxonomy provides a more unified view and framework from the theoretical grounding of the methods. Specifically, in our work, all contrastive methods are theoretically grounded by mutual information maximization, and the contrastive objectives are different upper bounds or estimators of mutual information. In our framework, different levels of contrast are determined by how views are selected or generated. We believe that the more unified view enables a more clear comparison and insightful understanding of commons and differences among SSL methods.}
    \item \revision{Though adapted to graphs, the taxonomy proposed by~\cite{liu2020self} is mostly oriented by SSL methods for images. While SSL for graphs and images share some connections and similarities, there are remarkable differences between specific methods for the two types of data and some categorizations of images do not apply to graphs. For example, the relative position and the cluster discrimination methods categorized by~\cite{liu2020self} are image-specific contrastive methods and do not apply to graphs whereas view generation approaches for graphs and graph-specific predictive tasks are not discussed in~\cite{liu2020self}. Therefore, graph-specific SSL reviews such as ours are important and necessary for a better understanding of existing methods and benefit future studies.}
\end{itemize}
\newrevision{Recently, another concurrent survey~\cite{liu2021graph} provides reviews on SSL methods for GNNs. The work proposes a taxonomy with more subdivided categories including generation-based, auxiliary property-based, contrastive-based, and hybrid methods. While~\cite{liu2021graph} aims to provide better coverage on existing SSL methods for review, our work focuses on providing a more timely and unified review under comparable frameworks and provide insights into future SSL studies.}

% \IEEEPARstart{T}{his}

% \begin{figure*}
%     \centering
%     \includegraphics[width=0.98\textwidth]{figures/methods_overview.pdf}
%     \caption{Caption}
%     \label{fig:my_label}
% \end{figure*}

\section{Problem Formulation}

\subsection{Notations}
%(Outline)

% 1. Notation for graph data (labeled and attributed), graph representation (node-level, graph-level)
We consider an attributed undirected graph $G=(V, E, \alpha)$, where $V=\{v_1,\cdots, v_{|V|}\}$ denotes the set of its nodes, $E=\{e_1\cdots, e_{|E|}\}$ denotes the set of its edges and $\alpha:V\to \mathbb{R}^d$ denotes the mapping from a node to its attributes of $d$ dimensions. We denote the adjacency matrix of $G$ by $\bm A\in\mathbb{R}^{|V|\times|V|}$, where $\bm A_{ij}=\mathbbm{1}\left[(v_i, v_j)\in E\right]$ for $1\le i,j\le |V|$, and denote the feature matrix of $G$ by $\bm X\in\mathbb{R}^{|V|\times d}$, where the $i$-th row $X_i=\alpha(v_i)$ for $1\le i\le |V|$. \revision{A heterogeneous graph additionally includes elements $\phi: V\to T_n$ that maps a node to a node type in $T_n$ and $\psi: E\to T_e$ that maps an edge to an edge type in $T_e$.}

% 2. Formulation of representation learning
Given the graph data $(\bm A, \bm X)$ from the input space $\mathcal{G}$, we are interested in the representation of the graph at either node-level or graph-level for any downstream prediction tasks. In general, we want to learn an encoder $f$ such that the representation $\bm H=f(\bm A, \bm X)$ can achieve desired performance on a downstream prediction task. For node-level prediction tasks, we learn a node-level encoder $f_n:\mathbb{R}^{|V|\times|V|}\times \mathbb{R}^{|V|\times d}\to \mathbb{R}^{|V|\times q}$ that takes the graph data $(\bm A, \bm X)$ as inputs and computes the representations for all nodes $\bm H_{node}=f_{node}(\bm A, \bm X)\in\mathbb{R}^{|V|\times q}$. For graph-level prediction tasks, we learn a graph-level encoder $f_g:\mathbb{R}^{|V|\times|V|}\times \mathbb{R}^{|V|\times d}\to \mathbb{R}^{q}$ that computes a single vector $\bm h_{graph}=f_{graph}(\bm A, \bm X)\in\mathbb{R}^{q}$ as the representation of the given graph. Practically, graph-level encoders are usually constructed as a node-level encoder followed by a readout function. In many cases, a model for node-level representation learning may also be utilized to compute graph-level representations by adding an appropriate readout function, and \emph{vice versa} (by removing the readout function).

% 3. Apply self-supervision
We let $\mathcal{P}$ denote the distribution of unlabeled graphs over the input space $\mathcal{G}$. Given a training dataset, the distribution $\mathcal{P}$ can be simply constructed as the uniform distribution over samples in the dataset. The self-supervision can contribute to the learning of graph encoders $f$ by utilizing information from $\mathcal{P}$ and minimizing a self-supervised loss $\mathcal{L}_{ssl}(f, \mathcal{P})$ determined by a specifically designed self-supervised learning task.

\subsection{Paradigms for Self-Supervised Learning}
Typical training paradigms to apply the self-supervision include unsupervised representation learning, unsupervised pretraining, and auxiliary learning. 

% 3. Three approaches to utilize self-supervision
In \textbf{unsupervised representation learning}, only the distribution $\mathcal{P}$ of unlabeled graphs is available for the entire training process. The problem of learning the representation of a given graph data $(\bm A, \bm X)\sim \mathcal{P}$ is formulated as
\begin{align}\label{unsupervised}
    f^* &= \arg\min_f \mathcal{L}_{ssl}(f, \mathcal{P}),\\
    \bm H^* &= f^*(\bm A, \bm X).
\end{align}
The learned representations $\bm H^*$ are then used in further downstream tasks such as linear classification and clustering.

The \textbf{unsupervised pretraining}, \revision{also dubbed the decoupled training by~\citet{wang2021decoupling}}, is usually performed as a two-stage training~\cite{jin2020self}. It first trains the encoder $f$ with unlabeled graphs. The pre-trained encoder $f_{init}$ is then used as the initialization of the encoder in a supervised fine-tuning stage. Formally, in addition to the distribution $\mathcal{P}$, the learner further gains access to a distribution $\mathcal{P}$ of labeled graphs over $\mathcal{G}\times\mathcal{Y}$, where $\mathcal{Y}$ denotes the label space. Again, given the labeled training dataset, $\mathcal{P}$ can be simply constructed as the uniform distribution over labeled samples in the dataset.
% The encoder $f$ is first optimized on $\mathcal{P}$ in an unsupervised fashion
% \begin{equation}
%     f^* = \arg\min_f \mathcal{L}_{ssl}(f, \mathcal{P}).
% \end{equation}
% in the pretraining stage.
The pretrained encoder $f_{init}$ is then fine-tuned on $\mathcal{P}$, together with a prediction head $h$ s.t. $h(f(\bm A, \bm X))\in \mathcal{Y}$ and a supervised loss $\mathcal{L}_{sup}$, i.e.,
\begin{equation}
    f^*, h^* = \arg\min_{(f,h)} \mathcal{L}_{sup}(f, h, \mathcal{P}),
\end{equation}
with initialization
\begin{equation}
    f_{init} = \arg\min_f \mathcal{L}_{ssl}(f, \mathcal{P}).
\end{equation}

The unsupervised pretraining and supervised finetuning paradigm is considered as the most strategy to perform semi-supervised learning and transfer learning. For semi-supervised learning, the labeled graphs in the finetuning dataset is a portion of the pretraining dataset. For transfer learning, the pretraining and finetuning datasets are from different domains. Note that a similar learning setting called unsupervised domain adaptation has also been studied generally~\cite{ganin2015unsupervised} or specifically in the image domain~\cite{sun2019unsupervised}, where the encoder is pre-trained on labeled data but finetuned on unlabeled data under self-supervision. Since the paradigm is not specifically studied in the graph domain, we do not discuss the learning setting in detail in this survey.

The \textbf{auxiliary learning}, also known as joint training~\cite{jin2020self}, aims to improve the performance of a supervised primary learning task by including an auxiliary task under self-supervision. Formally, we let $\mathcal{Q}$ denote the joint distribution of graph data and labels for the primary learning task and $\mathcal{P}$ denote the marginal of graph data. We want to learn both the decoder $f$ and the prediction head $h$, where $h$ is trained under supervision on $\mathcal{Q}$ and $f$ is trained under both supervision and self-supervision on $\mathcal{P}$. The learning problem is formulated as
\begin{equation}
    f^*, h^* = \arg\min_{(f,h)} \mathcal{L}_{sup}(f, h, \mathcal{Q}) + \lambda \mathcal{L}_{ssl}(f, \mathcal{P}),
\end{equation}
where $\lambda$ is a positive scalar weight that balances the two terms in the loss.

\begin{figure}
    \centering
    \includegraphics[width=0.48\textwidth]{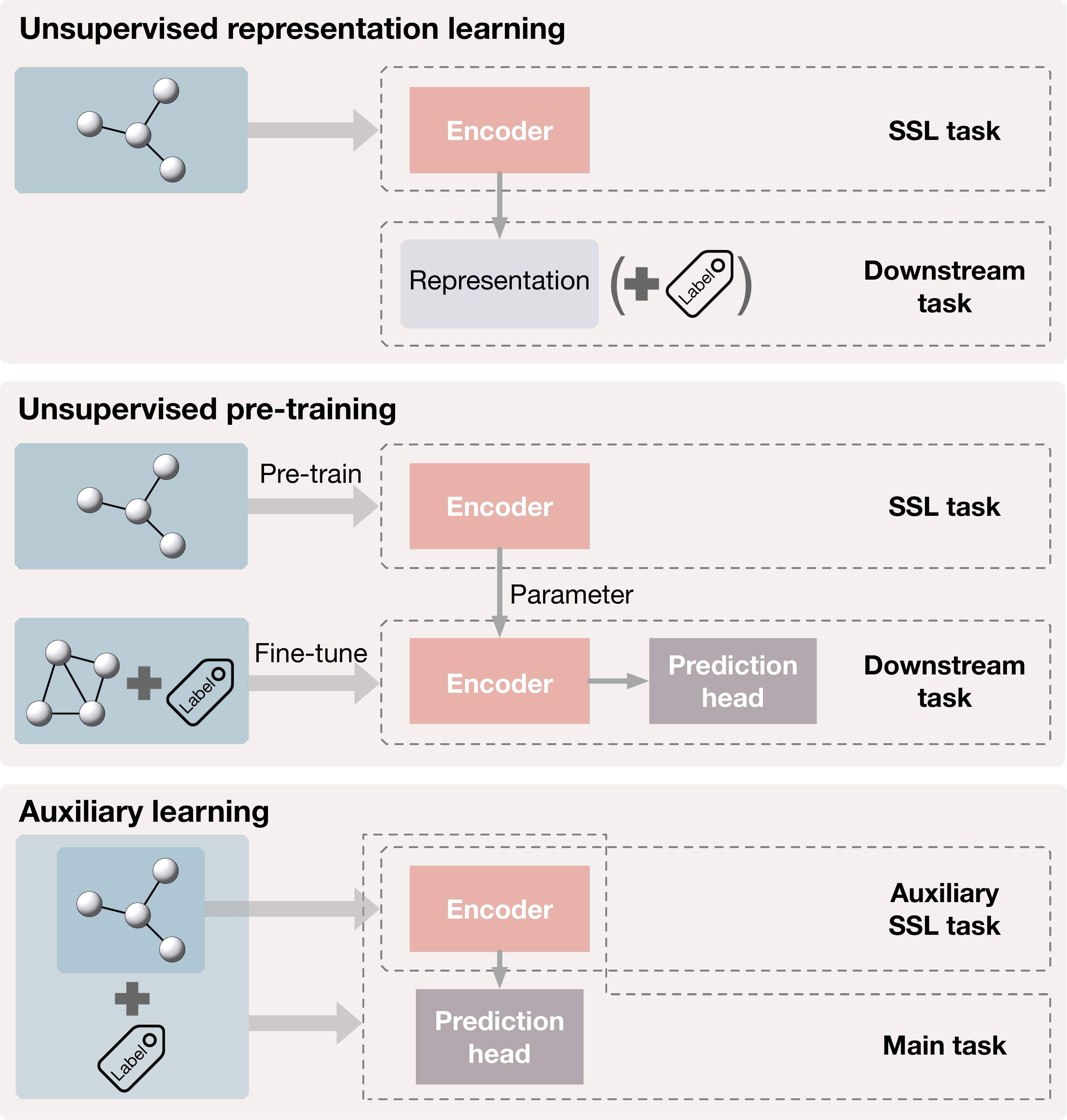}
    \caption{Paradigms for self-supervised learning. \textbf{Top}:in unsupervised representation learning, graphs only are used to train the encoder through the self-supervised task. The learned representations are fixed and used in downstream tasks such as linear classification and clustering. \textbf{Middle}: unsupervised pre-training trains the encoder with unlabeled graphs by the self-supervised task. The pre-trained encoder's parameters are then used as the initialization of the encoder used in supervised fine-tuning for downstream tasks. \textbf{Bottom}: in auxiliary learning, an auxiliary task with self-supervision is included to help learn the supervised main task. The encoder is trained through both the main task and the auxiliary task simultaneously.}
    \label{fig:paradigms}
    \vspace{-6pt}
\end{figure}

\section{Contrastive Learning}
The study of contrastive learning has made significant progress in natural language processing and computer vision. Inspired by the success of contrastive learning in images, recent studies propose similar contrastive frameworks to enable self-supervised training on graph data. Given training graphs, contrastive learning aims to learn one or more encoders such that representations of similar graph instances agree with each other, and that representations of dissimilar graph instances disagree with each other. We unify existing approaches to constructing contrastive learning tasks into a general framework that learns to discriminate jointly sampled view pairs (\emph{e.g.} two views belonging to the same instance) from independently sampled view pairs (\emph{e.g.} views belonging to different instances). In particular, we obtain multiple views from each graph in the training dataset by applying different transformations. Two views generated from the same instance are usually considered as a positive pair and two views generated from different instances are considered as a negative pair. The agreement is usually measured by metrics related to the mutual information between two representations.

One major difference among graph contrastive learning methods lies in (a) the objective for discrimination task given view representations. In addition, due to the unique data structure of graphs, graph contrastive learning methods also differ in (b) approaches that views are obtained, and (c) graph encoders that compute the representations of views. A graph contrastive learning method can be determined by specifying its components (a)--(c). In this section, we summarize graph contrastive learning methods in a unified framework and then introduce (a) and (b) individually used in existing studies. In Appendix~C, we introduce graph neural networks specifically adopted in contrastive learning as graph encoders and provide further comparisons and discussions on their effects in contrastive learning. Moreover, we summarize all contrastive methods being reviewed by this survey in Supplementary Table~1 for more clear comparisons.

\begin{figure*}
    \centering
    \includegraphics[width=0.98\textwidth]{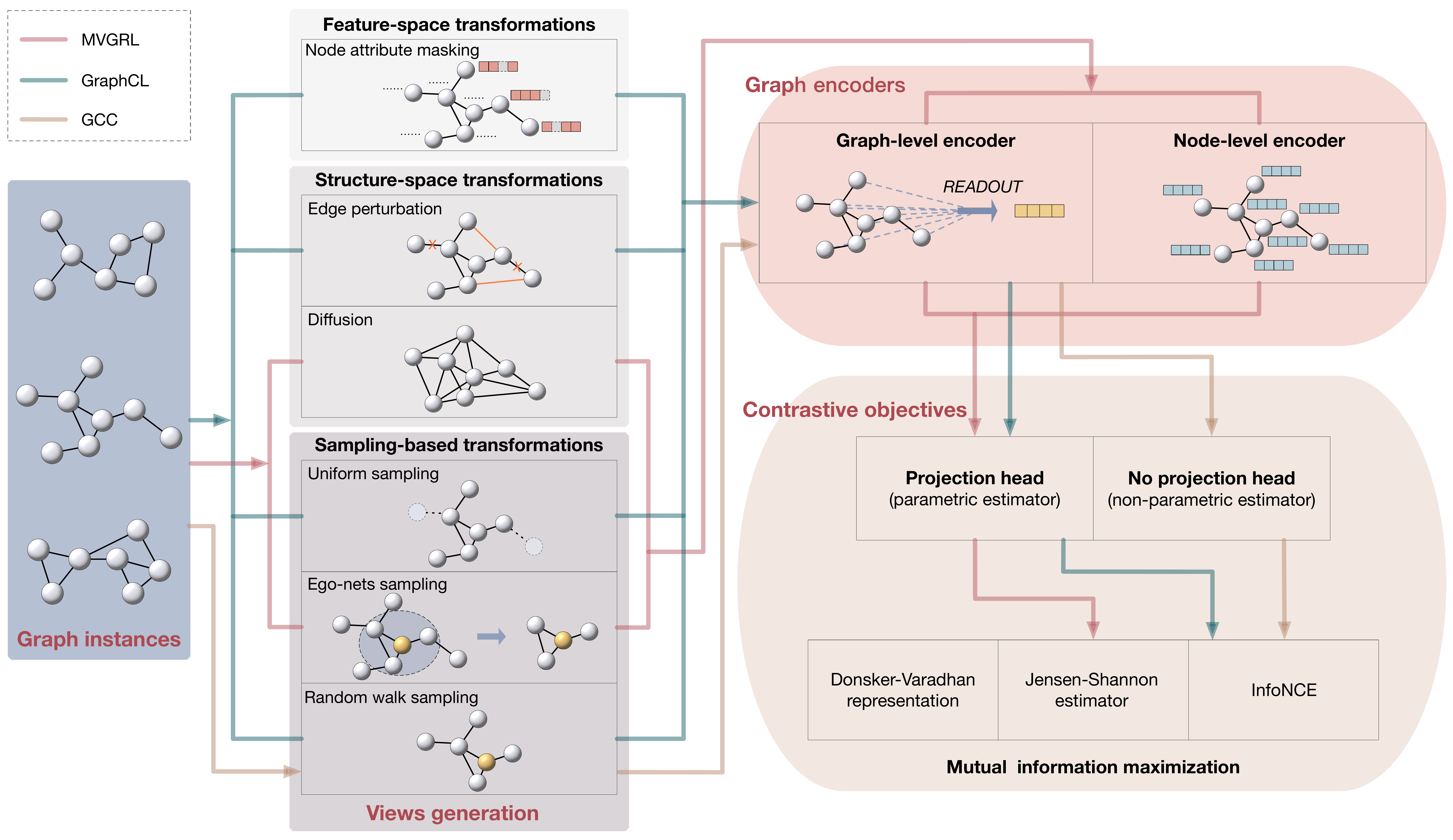}
    \caption{The general framework of contrastive methods. A contrastive method can be determined by defining its views generation, encoders, and objective. Different views can be generated by a single or a combination of instantiations of three types of transformations. Commonly employed transformations include node attribute masking as feature transformation, edge perturbation and diffusion as structure transformations, and uniform sampling, ego-nets sampling, and random walk sampling as sample-based transformations. Note that we consider a node representation in node-graph contrast~\cite{velikovi2019deep, sun2019infograph, tian2019contrastive} as a graph view with ego-nets sampling followed by a node-level encoder. For graph encoders, most methods employ graph-level encoders and node-level encoders are usually used in node-graph contrast. Common contrastive objectives include Donsker-Varadhan representation, Jensen-Shannon estimator, InfoNCE, and other non-bound objectives. An estimator is parametric if projection heads are employed, and is non-parametric otherwise. Examples of three specific contrastive learning methods are illustrated in the figure. Red lines connect options used in MVGRL~\cite{hassani2020contrastive}; green lines connect options adopted in GraphCL~\cite{you2020graph}; yellow lines connect options taken in GCC~\cite{qiu2020gcc}.}
    \label{fig:contrastive}
\end{figure*}

\subsection{Overview of Contrastive Learning Framework}
In general, key components that specify a contrastive learning framework include transformations that compute multiple views from each given graph, encoders that compute the representation for each view, and the learning objective to optimize parameters in encoders. An overview of the framework is shown in Figure~\ref{fig:contrastive}. Concretely, given a graph $(\bm A,\bm X)$ as a random variable distributed from $\mathcal{P}$, multiple transformations $\mathcal{T}_1,\cdots,\mathcal{T}_k$ are applied to obtain different views $\bm w_1,\cdots,\bm w_k$ of the graph. Then, a set of encoding networks $f_1,\cdots,f_k$ take corresponding views as their inputs and output the representations $\bm h_1,\cdots,\bm h_k$ of the graph from each views. Formally, we have
\begin{align}
    \bm w_i &= \mathcal{T}_i(\bm A,\bm X),\\
    \bm h_i &= f_i(\bm w_i),\; i=1,\cdots,k.
\end{align}
We assume $\bm w_i = (\hat{\bm A}_i,\hat{\bm X}_i) = \mathcal{T}_i(\bm A,\bm X)$ in this survey since existing contrastive methods consider their views as graphs. However, note that not all views $w_i$ are necessarily graphs or sub-graphs in a general sense. In addition, certain encoders can be identical to each other or share their weights.

During training, the contrastive objective aims to train encoders to maximize the agreement between view representations computed from the same graph instance. The agreement is usually measured by the mutual information $\mathcal{I}(\bm h_i, \bm h_j)$ between a pair of representations $\bm h_i$ and $\bm h_j$. We formalize the contrastive objective as
\begin{align}\label{eq:objective}
    \max_{\{f_i\}_{i=1}^k} \frac{1}{\sum_{i\ne j} \sigma_{ij}}\left[
    \sum_{i\ne j} \sigma_{ij}\mathcal{I}(\bm h_i, \bm h_j)\right],
\end{align}
where $\sigma_{ij}\in\{0,1\}$, and $\sigma_{ij}=1$ if the mutual information is computed between  $\bm h_i$ and $\bm h_j$, and $\sigma_{ij}=0$ otherwise, and $\bm h_i$ and $\bm h_j$ are considered as two random variables belonging to either a joint distribution or the product of two marginals. To enable efficient computation of the mutual information, certain estimators $\mathcal{\widehat I}$ of the mutual information are usually used instead as the learning objective. Note that some contrastive methods apply projection heads~\cite{you2020graph,hassani2020contrastive} to the representations. For the sake of uniformity, we consider such projection heads as parts of the computation in the mutual information estimation.

\begin{figure}[t]
    \centering
    \includegraphics[width=0.48\textwidth]{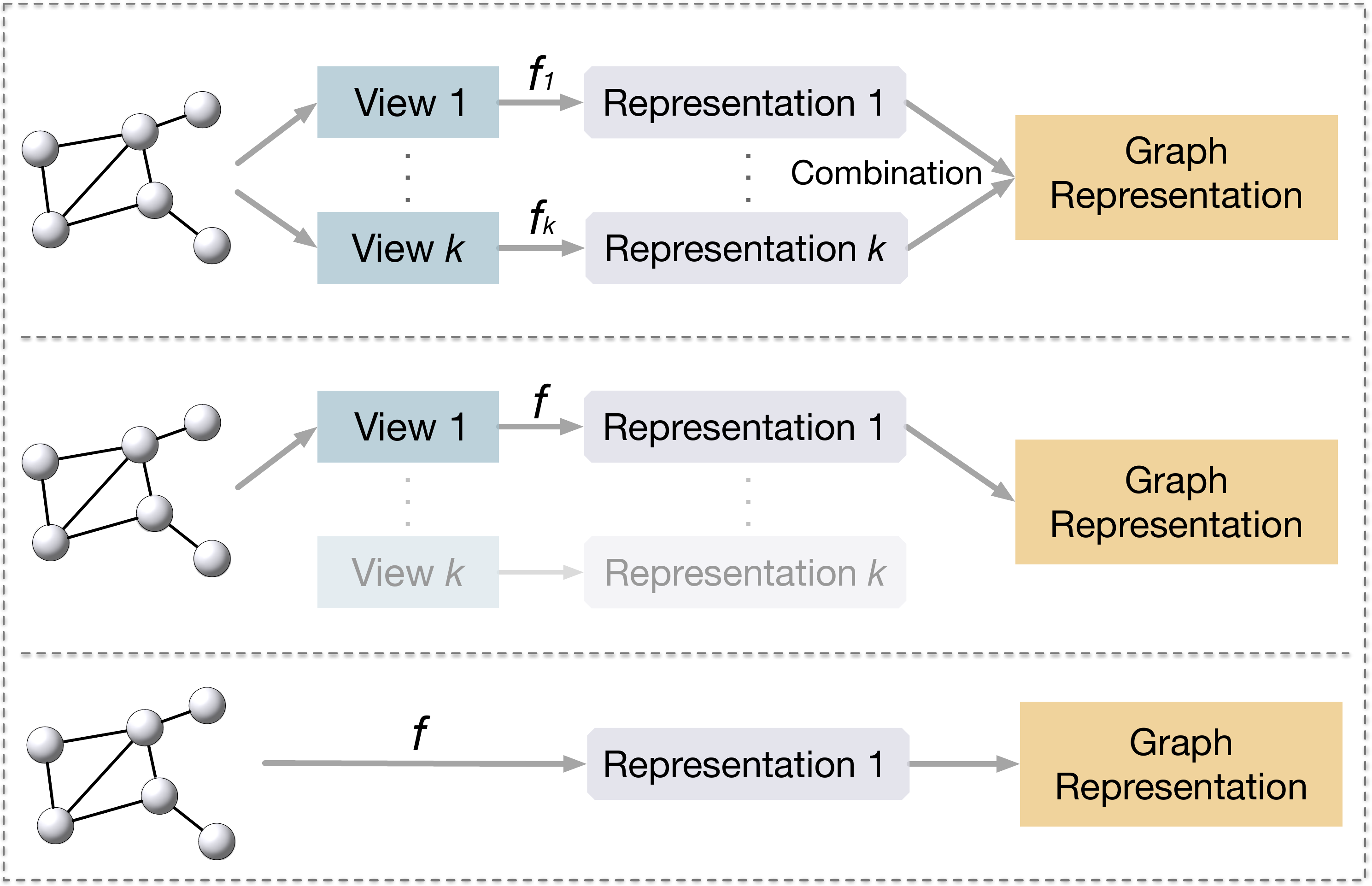}
    \caption{Different ways of using encoders during inference. \textbf{Top}: encoders for multiple views are used and output representations are merged by combinations such as summation~\cite{hassani2020contrastive} or concatenation. \textbf{Middle}: only the main encoder~\cite{thakoor2021bootstrapped} and the corresponding view are used during inference. \textbf{Bottom}: the given graph is directly input to the only encoder~\cite{you2020graph, qiu2020gcc} shared by all views to compute its representation.}
    \label{fig:enc_infer}
    \vspace{-8pt}
\end{figure}

During inference, one can either use a single trained encoder to compute the representation or a combination of multiple view representations such as the linear combination or the concatenation as the final representation of a given graph. Three examples of using encoders in different ways during inference are illustrated in Figure~\ref{fig:enc_infer}.

\subsection{Contrastive Objectives based on MI Estimations}

% \subsubsection{}
Given a pair of random variables $(\bm x, \bm y)$, the mutual information $\mathcal{I}(\bm x, \bm y)$ measures the information that $\bm x$ and $\bm y$ share, given by
\begin{align}
    \mathcal{I}(\bm x, \bm y)
    &= D_{KL} (p(\bm x, \bm y)||p(\bm x)p(\bm y))\\
    &= \mathbb{E}_{p(\bm x, \bm y)}\left[\log\frac{p(\bm x,\bm y)}{p(\bm x)p(\bm y)}\right],
    % &= \mathbb{E}_{p(\bm x, \bm y)}\left[\log\frac{p(\bm x|\bm y)}{p(\bm x)}\right],
\end{align}
where $D_{KL}$ denotes the Kullback-Leibler (KL) divergence. The contrastive learning seeks to maximize the mutual information between two views as two random variables. In particular, it trains the encoders to be contrastive between representations of a positive pair of views that comes from the joint distribution $p(\bm v_i, \bm v_j)$ and representations of a negative pair of views that comes from the product of marginals $p(\bm v_i)p(\bm v_j)$.

In order to computationally estimate and maximize the mutual information in the contrastive learning, three typical lower-bounds to the mutual information are derived~\cite{hjelm2018learning}, namely, the Donsker-Varadhan representation $\mathcal{\widehat I}^{(DV)}$~\cite{donsker1983Asymptotic,belghazi2018mutual}, the Jensen-Shannon estimator $\mathcal{\widehat I}^{(JS)}$~\cite{nowozin2016f} and the noise-contrastive estimation $\mathcal{\widehat I}^{(NCE)}$ (InfoNCE)~\cite{gutmann2010noise, oord2018representation}. \revision{Among the three lower-bounds, $\mathcal{\widehat I}^{(JS)}$ and $\mathcal{\widehat I}^{(NCE)}$ are commonly used as objectives~\cite{you2020graph, hassani2020contrastive, velikovi2019deep, sun2019infograph} in the contrastive learning in graphs.}

A mutual information estimation is usually computed based on a discriminator $\mathcal{D}:\mathbb{R}^q\times\mathbb{R}^q\to\mathbb{R}$ that maps the representations of two views to an agreement score between the two representations. The discriminator $\mathcal{D}$ can be either parametric or non-parametric. For example, the discriminator can optionally apply a set of projection heads~\cite{you2020graph,hassani2020contrastive} to the representations $\bm h_1,\cdots,\bm h_k$ before computing the pair-wise similarity. We formalize the optional projection heads as $g_1, \cdots, g_k$ such that
\begin{equation}
    \bm z_i = g_i(\bm h_i),\; i=1,\cdots,k,
\end{equation}
where $g_i$ can be an identical mapping, a linear projection or an MLP. Parameterized $g_i$ are optimized simultaneously with the encoders $f_i$ in Eqn.~(\ref{eq:objective}), given by
\begin{align}
    \max_{\{f_i, g_i\}_{i=1}^k} \frac{1}{\sum_{i\ne j} \sigma_{ij}}\left[
    \sum_{i\ne j} \sigma_{ij}\mathcal{\widehat I}_{g_i, g_j}(\bm h_i, \bm h_j)\right],
\end{align}

In following subsections, we introduce the three lower bounds as specific estimations of mutual information and a non-bound estimation of mutual information. We further compare and discuss the effect of different MI estimations in contrastive learning in Appendix~D.

\subsubsection{Donsker-Varadhan Estimator}
The Donsker-Varadhan (DV) estimator, also knwon as the DV representation of the KL divergence, is a lower-bound to the mutual information and hence can be applied to maximize the mutual information. Given $\bm h_i$ and $\bm h_j$, the lower-bound is computed as
\begin{equation}\label{eq:dv0}
\begin{split}
    \mathcal{\widehat I}^{(DV)}(\bm h_i, \bm h_j)
    = \mathbb{E}_{p(\bm h_i, \bm h_j)}&[\mathcal{D}(\bm h_i, \bm h_j)] \\
    - & \log\mathbb{E}_{p(\bm h_i)p(\bm h_j)}\left[e^{\mathcal{D}(\bm h_i, \bm h_j)}\right],
\end{split}
\end{equation}
where $p(\bm h_i, \bm h_j)$ denotes the joint distribution of the two representations $\bm h_i, \bm h_j$ and $p(\bm h_i)p(\bm h_j)$ denotes the product of marginals. For simplicity and to include the graph data distribution $\mathcal{P}$, we assume transformations $\mathcal{T}_i$ to be deterministic and encoders $f_i$ to be injective, and have $p(\bm h_i, \bm h_j)=p(\bm h_i)p(\bm h_j|\bm h_i)=p\big(f_i(\mathcal{T}_i(\bm A, \bm X))\big)p\big(f_j(\mathcal{T}_j(\bm A, \bm X))\big|(\bm A, \bm X))$. We hence re-write Eqn.~(\ref{eq:dv0}) as
\begin{equation}\label{eq:dv}
\begin{split}
    \mathcal{\widehat I}^{(DV)}(\bm h_i, \bm h_j)
    = \mathbb{E}_{(\bm A, \bm X)\sim\mathcal{P}}[\mathcal{D}(\bm h_i, \bm h_j)]& \\
    - \log\mathbb{E}_{[(\bm A, \bm X),(\bm A', \bm X')]\sim\mathcal{P}\times\mathcal{P}}&\left[e^{\mathcal{D}(\bm h_i, \bm h_j')}\right],
\end{split}
\end{equation}
where $\bm h_i$ and $\bm h_j$ in the first term are computed from $(\bm A, \bm X)$ distributed from $\mathcal{P}$, $\bm h_i$ and $\bm h_j'$ in the second term are computed from $(\bm A, \bm X)$ and $(\bm A', \bm X')$ identically and independently distributed from $\mathcal{P}$, respectively. In following formulations, we use the later notation includes $\mathcal{P}$.

\subsubsection{Jensen-Shannon Estimator}
Compared to the Donsker-Varadhan estimator, the Jensen-Shannon (JS) estimator enables more efficient estimation and optimization of the mutual information by computing the JS-divergence between the joint distribution and the product of marginals.

Given two representations $\bm h_i$ and $\bm h_j$ computed from the random variable $(\bm A, \bm X)$ and a discriminator $\mathcal{D}$, \textbf{DGI}~\cite{velikovi2019deep}, \textbf{InfoGraph}~\cite{sun2019infograph}, \citet{Hu2020Strategies} and \textbf{MVGRL}~\cite{hassani2020contrastive} computes the JS estimator
\begin{equation}\label{eq:js}
\begin{split}
    \mathcal{\widehat I}^{(JS)}(\bm h_i, &\bm h_j) = \mathbb{E}_{(\bm A, \bm X)\sim\mathcal{P}}\left[\log(\mathcal{D}(\bm h_i, \bm h_j))\right] + \\
    &\mathbb{E}_{[(\bm A, \bm X),(\bm A', \bm X')]\sim\mathcal{P}\times\mathcal{P}}\left[\log(1-\mathcal{D}(\bm h_i, \bm h_j'))\right],
\end{split}
\end{equation}
where $\bm h_i, \bm h_j$ in the first term are computed from $(\bm A, \bm X)$ distributed from $\mathcal{P}$, $\bm h_i$ and $\bm h_j'$ in the second term are computed from $(\bm A, \bm X)$ and $(\bm A', \bm X')$ identically and independently distributed from the distribution $\mathcal{P}$. To further include edge features, \citet{peng2020graph} derives the graphic mutual information (\textbf{GMI}) based on MI decomposition and optimizes GMI via JS estimator. Note that \cite{sun2019infograph} and \cite{hassani2020contrastive} depict a softplus version of the JS estimator,
\begin{equation}\label{eq:js-sp}
\begin{split}
    \mathcal{\widehat I}^{(JS-SP)}&(\bm h_i, \bm h_h) = \mathbb{E}_{(\bm A, \bm X)\sim\mathcal{P}}\left[-sp(-\mathcal{D}'(\bm h_i, \bm h_j))\right] - \\
    &\mathbb{E}_{[(\bm A, \bm X),(\bm A', \bm X')]\sim\mathcal{P}\times\mathcal{P}}\left[sp(\mathcal{D}'(\bm h_i, \bm h_j'))\right],
\end{split}
\end{equation}
where $sp(x) = \log(1+e^x)$. We consider the JS estimators in Eqn.~(\ref{eq:js}) and Eqn.~(\ref{eq:js-sp}) to be equivalent by letting $\mathcal{D}(\bm h_i, \bm h_j)=\mbox{sigmoid}(\mathcal{D}'(\bm h_i, \bm h_j))$.

For the the negative pairs of graphs $[(\bm A, \bm X),(\bm A', \bm X')]\sim\mathcal{P}\times\mathcal{P}$ in particular, \textbf{DGI}~\cite{velikovi2019deep} samples one graph $(\bm A, \bm X)$ from the training dataset and applies a stochastic corruption $\mathcal{C}$ to obtain $(\bm A', \bm X')=\mathcal{C}(\bm A, \bm X)$. For node-level tasks, \textbf{MVGRL}~\cite{hassani2020contrastive} follows \textbf{DGI} to obtain negative samples by corrupting given graphs. \textbf{InfoGraph}~\cite{sun2019infograph} independently samples two graphs from the training dataset as a negative pair, which is followed by \textbf{MVGRL} for graph-level tasks. Discriminators in JS estimators usually compute the agreement score between two vectors by their inner product with sigmoid, \emph{i.e.}, $\mathcal{D}(\bm h_i, \bm h_j)=\mbox{sigmoid}(\bm z_i^T\bm z_j)=\mbox{sigmoid}(g_i(\bm h_i)^Tg_j(\bm h_j))$.

% For the projection heads $g_i$, \textbf{DGI}~\cite{velikovi2019deep} uses identical mapping. \textbf{InfoGraph}~\cite{sun2019infograph} constructs the projection heads $g_i$ as 3-layer MLPs with ReLU nonlinearity and jumping knowledge. For \cite{hassani2020contrastive}, the projection heads $g_i$ are MLPs with 2 hidden layers and PReLU nonlinearity. All three methods computes the dot product between the (projected) representations $\mathcal{D}(\bm h_i, \bm h_j)=\mbox{sigmoid}(\bm z_i^T\bm z_j)=\mbox{sigmoid}(g_i(\bm h_i)^Tg_j(\bm h_j))$.

\subsubsection{InfoNCE}
InfoNCE $\mathcal{\widehat I}^{(\mathrm{NCE})}$ is another lower-bound to the mutual information $\mathcal{I}$. It is shown by~\citet{you2020graph} that maximizing InfoNCE it equivalent to maximizing the Donsker-Varadhan estimator. Given the representations $\bm h_i$ and $\bm h_j$ of two views of random variable $(\bm A, \bm X)$, the discriminator $\mathcal{D}$, and the number of negative samples $N$, the InfoNCE is formalized as
\begin{equation}\label{eq:infonce}
\begin{split}
    \mathcal{\widehat I}^{(\mathrm{NCE})}(\bm h_i, \bm h_j) = \mathbb{E}_{(\bm A, \bm X)\sim\mathcal{P}}\Bigg[\mathcal{D}(\bm h_i, \bm h_j) - \quad\quad\quad&\\
    \mathbb{E}_{\bm K\sim\mathcal{P}^{N}}\Bigg[\log\sum_{(\bm A', \bm X')\in\bm K}e^{\mathcal{D}(\bm h_i, \bm h_j')}/N\Bigg|(\bm A, \bm X)\Bigg]\Bigg]&\\
    = \mathbb{E}_{[(\bm A, \bm X),\bm K]\sim\mathcal{P}\times\mathcal{P}^{N}}\left[\log\frac{e^{\mathcal{D}(\bm h_i, \bm h_j)}}{\sum_{(\bm A', \bm X')\in\bm K}e^{\mathcal{D}(\bm h_i, \bm h_j')}}\right]&\\
    +\log N&,
\end{split}
\end{equation}
where $\bm K$ consists of $N$ random variables identically and independently distributed from $\mathcal{P}$, $\bm h_i, \bm h_j$ are the representations of the $i$-th and $j$-th views of $(\bm A, \bm X)$, and $\bm h_j'$ is the representation of the $j$-th view of $(\bm A', \bm X')$.

In practice, we compute the InfoNCE on mini-batches of size $N+1$. For each sample $\bm x$ in a mini-batch $\bm B$, we consider the set of the rest $N$ samples as a sample of $\bm K$. We then discard the constant term $\log N$ in Eqn.~(\ref{eq:infonce}) and minimize the loss
\begin{equation}
    \mathcal{L}_{\mathrm{InfoNCE}} =
    -\frac{1}{N+1}\sum_{\bm x\in\bm B}
    \left[\log\frac{e^{\mathcal{D}(\bm h_i, \bm h_j)}}{\sum_{\bm x'\in\bm B\setminus\{\bm x\}}e^{\mathcal{D}(\bm h_i, \bm h_j')}}\right].
\end{equation}
Intuitively, the optimization of InfoNCE loss aims to score the agreement between $\bm h_i$ and $\bm h_j$ of views from the same instance $\bm x$ higher than between $\bm h_i$ and $\bm h_j'$ from the rest $N$ negative samples $\bm B\setminus\{\bm x\}$. \textbf{GraphCL}~\cite{you2020graph} and \textbf{GRACE}~\cite{zhu2020deep} include additional contrast among the same view of different instances (\emph{i.e.}, $\bm h_i$ and $\bm h'_i$) or different nodes in the same view (intra-view contrast~\cite{zhu2020deep}), leading to the optimization of lower bounds of InfoNCE, which are still lower bounds of MI.

Discriminators in typical InfoNCE compute the agreement score between two vectors by their inner product, \emph{i.e.}, $\mathcal{D}(\bm h_i,\bm h_j)=\bm z_i^T\bm z_j=g_i(\bm h_i)^Tg_j(\bm h_j)$. A specific type of the InfoNCE loss, known as the \textbf{NT-Xent}~\cite{sohn2016improved} loss, includes normalization and a preset temperature parameter $\tau$ in the computation of discriminator $\mathcal{D}$ in the InfoNCE loss, \emph{i.e.}, $\mathcal{D}(\bm h_i, \bm h_j)=g_i(\bm h_i)^Tg_j(\bm h_j)/\tau$. The discriminator in \citet{you2020graph} computes the agreement score between vectors with normalizations, \emph{i.e.}, $\mathcal{D}(\bm h_i, \bm h_j)=\frac{g_i(\bm h_i)^Tg_j(\bm h_j)/\tau}{\norm{g_i(\bm h_i)}\norm{g_j(\bm h_j)}}$, where $\norm{\cdot}$ denotes the $\ell_2$-norm.

% For the discriminator $\mathcal{D}$ in particular, \cite{you2020graph} applies a 2-layer MLP as the shared projection head for all views and computes the score as $\mathcal{D}(\bm h_i, \bm h_j)=\frac{g_i(\bm h_i)^Tg_j(\bm h_j)/\tau}{\norm{g_i(\bm h_i)}\norm{g_j(\bm h_j)}}$, where $\tau$ is a preset temperature parameter and $\norm{\cdot}$ is the $l^2$-norm. In addition, \cite{qiu2020gcc} applies no projection heads with $g_i$ being identical mappings and computes the score as $\mathcal{D}(\bm h_i, \bm h_j)=g_i(\bm h_i)^Tg_j(\bm h_j)/\tau=\bm h_i^T\bm h_j/\tau$.

\subsubsection{Other Mutual Information Estimators}
There are other objectives that have been used in some studies, and optimizing these objectives can also encourage higher mutual information. Although the objectives differ from the above upper bound MI estimators

However, these objectives may not be provable lower-bounds to the mutual information, and optimizing these objectives does not guarantee the maximization of the mutual information.

For example, \citet{jiao2020sub} proposes to minimize the \textbf{triplet margin loss}~\cite{schroff2015facenet}, which is commonly used in deep metric learning~\cite{hoffer2015deep}. Given representations $\bm h_i$, $\bm h_j$ and the discriminator $\mathcal{D}$, the triplet margin loss is formalized as
\begin{equation}
\begin{split}
    \mathcal{L}_{\mathrm{triplet}}=\mathbb{E}_{[(\bm A, \bm X), (\bm A', \bm X')]\sim\mathcal{P}\times\mathcal{P}}&\big[
    \max\{\mathcal{D}(\bm h_i, \bm h_j)\\
    &- \mathcal{D}(\bm h_i, \bm h_j')+\epsilon,0\}
    \big],
\end{split}
\end{equation}
where $\mathcal{D}(\bm h_i, \bm h_j)$ is computed as $\mbox{sigmoid}(\bm h_i^T\bm h_j)$ or based on the Euclidean distance $\norm{\bm h_i - \bm h_j}$ and $\epsilon$ is the margin value. \revision{While the triplet loss differs from previous MI-based objectives in formulations, \citet{khosla2020supervised} show that the triplet loss is a special case of the InfoNCE (NT-Xent) loss when there is only one negative sample where the margin value $\epsilon$ corresponds to the temperature parameter $\tau$ in NT-Xent.} Moreover, the \textbf{Bayesian Personalized Ranking} (BPR) loss~\cite{rendle2009bpr} used in~\citet{jiao2020sub} is also equivalent to the InfoNCE loss when letting $N=1$ and $\mathcal{D}(\bm h_i,\bm h_j)=\bm h_i^T\bm h_j$.

\subsubsection{Projection Heads: Parametric MI Estimation}
Many contrastive learning studies~\cite{sun2019infograph, hassani2020contrastive, you2020graph} propose to include projection heads $g_i$ when computing the MI estimations. For example, \citet{sun2019infograph, hassani2020contrastive} use 3-layer MLPs, \citet{you2020graph} use 2-layer MLPs as the projection heads and \citet{sun2019infograph} applies a linear projection to the graph-level representation. The projection heads are shown to significantly improve the contrastive learning performance~\cite{chen2020simple}. \revision{For contrastive learning on heterogeneous graphs, it is common to apply individual projections to representations of different type of nodes. For example, \citet{jiang2021pre} adopt $\mathcal{D}(\bm h_u, \bm h_v) = [\bm W_{\phi(u)}\bm h_u]^T[\bm W_{\phi(v)}\bm h_v]=\bm h_u^T\bm W_R\bm h_v$ for nodes $u$ and $v$ with types $\phi(u)$ and $\phi(v)$ connected by the relationship $R$, where $\bm W_R=\bm W_{\phi(u)}^T\bm W_{\phi(b)}$.}

We consider MI estimators that include projection heads as parametric estimators and those without projection heads as non-parametric estimators. Then a reasonable explanation to the observation that the contrastive methods with projection heads usually achieve better performance is that parametric estimators provide better estimation to the mutual information.

\subsection{Graph View Generation}

To generate views from a graph sample distributed from $\mathcal{P}$, one usually applies different types of graph transformations (or augmentations) $\mathcal{T}$. Here, we only consider cases where $\mathcal{T}$ still outputs the graph-structured data. We summarize the existing transformations applied to graph data in three categories, feature transformations, structure transformations and sampling-based transformations. Feature transformations can be formalized as
\begin{equation}
    \mathcal{T}_{\mathrm{feat}}(\bm A, \bm X) = (\bm A, \mathcal{T}_X(\bm X)),
\end{equation}
where $\mathcal{T}_X:\mathbb{R}^{|V|\times d}\to\mathbb{R}^{|V|\times d}$ performs the transformation on the feature matrix $\bm X$. Structure transformations can be formalized as
\begin{equation}
    \mathcal{T}_{\mathrm{struct}}(\bm A, \bm X) = (\mathcal{T}_A(\bm A), \bm X),
\end{equation}
where $\mathcal{T}_A:\mathbb{R}^{|V|\times |V|}\to\mathbb{R}^{|V|\times |V|}$ performs the transformation on the adjacency matrix $\bm A$. And the sampling-based transformations are in the form
\begin{equation}
    \mathcal{T}_{\mathrm{sample}}(\bm A, \bm X) = (\bm A[S;S], \bm X[S]),
\end{equation}
where $S\subseteq V$ denotes a subset of nodes and $[\cdot]$ selects certain rows (and columns) from a matrix based on indices of nodes in $S$. We consider the transformations applied in the existing contrastive learning methods to generate different views as a single or a combination of instantiations of the three types of transformations above. Note that when node-level representations are of interest, the node-level contrasts are usually included. We consider nodes representations to be computed from views generated by ego-nets sampling from given graphs.

\subsubsection{Feature Transformations}
Given an input graph $(\bm A, \bm X)$, a feature transformation only performs transformation to the attribute matrix $\bm X$, \emph{i.e.}, $\mathcal{T}(\bm A, \bm X) = (\bm A, \mathcal{T}_X(\bm X))$.

\textbf{Node attribute masking}~\cite{you2020graph} is one of the most common way to apply the feature transformations. The node attribute masking randomly masks a small portion of attributes of all node with constant or random values. Concretely, given the input attribute matrix $\bm X$, we specify $\mathcal{T}_X(\bm X)$ for the node attribute masking as
\begin{equation}\label{eq:masking}
    \mathcal{T}_X^{(mask)}(\bm X) = \bm X*(\bm1 - \bm1_m) + \bm M*\bm1_m,
\end{equation}
where $*$ denotes the element-wise multiplication, $\bm M$ denotes a matrix with masking values and $\bm{1}_m$ denotes the masking location indicator matrix. Given the masking ratio $r$, elements in $\bm{1}_m$ are set to $1$ individually with a probability $r$ and $0$ with a probability $1-r$. To employ adaptive masking, \citet{zhu2021graph} propose to sample $\bm{1}_m$ with centrality-based probabilities, including degree centrality, eigenvector centrality, and PageRank centrality. The values in $\bm M$ specifies different masking strategies. For example, $\bm M=\bm 0$ applies a constant masking, $\bm M\sim N(\bm 0,\bm\Sigma)$ replaces the original values by Gaussian noise and $\bm M\sim N(\bm X,\bm\Sigma)$ adds Gaussian noise to the original values. 

In addition to contrastive models such as~\cite{you2020graph}, attributes masking is also commonly applied in predictive models~\cite{xie2020noise2same,batson2019noise2self} for regularized reconstruction. The node attribute masking forces the encoders to captures better dependencies between the masked attributes and unmasked context attributes and recover the masked value from its context during encoding.

% list each type of augmentation using /paragraph{}, include
    % the description to the augmentation
    % insights of the augmentation
    % cite the related paper

% optional: possible future directions

\subsubsection{Structure Transformations}
Given an input graph $(\bm A, \bm X)$, a structure transformation only performs transformation to the adjacency matrix $\bm A$ and remains $\bm X$ to be the same, \emph{i.e.}, $\mathcal{T}(\bm A, \bm X) = (\mathcal{T}_A(\bm A), \bm X)$. Existing contrastive methods apply two types of structure transformations, edge perturbation that randomly adds or drops edges between pairs of nodes and graph diffusion that creates new edges based on the accessibility from one node to another.

\textbf{Edge perturbation}~\cite{you2020graph,qiu2020gcc} randomly adds or drops edges in a given graph. Similarly to the node attribute masking, it applies masks to the adjacency matrix $\bm A$. In particular, we have
\begin{equation}
    \mathcal{T}_A^{(pert)}(\bm A) = \bm A*(\bm1 - \bm1_p) + (\bm1 - \bm A)*\bm1_p,
\end{equation}
where $*$ denotes the element-wise multiplication and $\bm1_p$ denotes the perturbation location indicator matrix. Given the perturbation ratio $r$, elements in $\bm{1}_p$ are set to $1$ individually with a probability $r$ and $0$ with a probability $1-r$. In addition, $\bm{1}_p$ is a symmetric matrix.

\textbf{Diffusion}~\cite{hassani2020contrastive} creates new connections between nodes based on random walks, aiming at generating a global view ($\bm S, \bm X$) of the graph in contrast to the local view ($\bm A, \bm X$). Two instantiations of diffusion transformations are proposed to use in~\cite{hassani2020contrastive}, namely, the heat kernel $\mathcal{T}_A^{(heat)}$ and the Personalized PageRank $\mathcal{T}_A^{(PPR)}$, formulated as follows.
\begin{equation}
    \mathcal{T}_A^{(heat)}(\bm A) = exp(t\bm A\bm D^{-1}-t),
\end{equation}
\begin{equation}
    \mathcal{T}_A^{(PPR)}(\bm A) = \alpha\left(\bm I_n - (1-\alpha)\bm D^{-1/2}\bm A\bm D^{-1/2}\right)^{-1},
\end{equation}
where $\bm D\in\mathbb{R}^{|V|\times|V|}$ is a diagonal degree matrix, $\alpha$ denotes the teleport probability in a random walk and $t$ denotes the diffusion time.

\textbf{Centrality-based edge removal}~\cite{zhu2021graph} randomly removes edges based on pre-computed probabilities determined by the centrality score of each edge. Centrality-based probabilities for edge removal reflects the importance of each edge, where less important edges are more likely to be removed. In particular, the centrality score of an edge $(u, v)\in E$ is computed as $w_{uv}=(\phi_c(u)+\phi_c(v))/2$, where $\phi_c(u)$ and $\phi_c(v)$ are the centrality of nodes $u$ and $v$ connected by the edge and a higher centrality score leads to lower probability $p_{uv}$ of edge removal.

\subsubsection{Sampling-Based Transformations}
We consider sampling-based transformations that sample node-induced sub-graphs from a given graph $(\bm A, \bm X)$, \emph{i.e.}, $\mathcal{T}_{\mathrm{sample}}(\bm A, \bm X) = (\bm A[S;S], \bm X[S])$ with $S\subseteq V$. Note that more generalized sub-graph sampling methods that sample both nodes from $V$ and edges from $E$ can be considered as a combination of the node-induced sub-graph sampling and the edge perturbation. As different sampling-based transformations are determined by the set $S$ of sampled nodes from the node-set $V$, we categorize the sampling-based transformations by how the set $S$ is obtained. Existing contrastive methods apply three approaches to obtain the node subset $S$, uniform sampling, random walk sampling, and ego-nets sampling.

\textbf{Uniform sampling} and \textbf{nodes dropping} can be considered as the two simplest sampling-based transformation approaches. The transformation in~\cite{hassani2020contrastive} samples sub-graphs by uniformly sampling a given number of nodes $S$ from $V$ and edges of the sampled nodes. In addition,
transformation methods in~\cite{you2020graph} include node dropping as one of the graph transformations, where each node has a certain probability to be dropped from the graph. We denote the set of dropped nodes by $D$ and we have $S = V\setminus D$.

\textbf{Ego-nets sampling} can be considered as a sampling-based transformation to unify the contrast performed between graph-level representation and node-level representations in the general contrastive learning framework, such as in \textbf{DGI}~\cite{velikovi2019deep}, \textbf{InfoGraph}~\cite{sun2019infograph} and \textbf{MVGRL}~\cite{hassani2020contrastive}. In other words, we consider that node-level representations are computed by node-level encoders from certain views, namely, ego-nets, of a given graph. Given a typical graph encoder with $L$ layers, the computation of the representation of each node $v_i$ only depends on its $L$-hop neighborhood, also known as the $L$-ego-net of node $v_i$. We hence consider the computing of node-level representations as performing $L$-ego-net sampling and a node-level encoder with $L$ layers. In particular, for each node $v_i$ in a given graph, the transformation $\mathcal{T}_i$ samples the $L$-ego net surrounding node $v_i$ as the view $w_i$, computed as
\begin{equation}
\bm w_i = \mathcal{T}_i(\bm A, \bm X)= (\bm A[\mathcal{N}_L(v_i);\mathcal{N}_L(v_i)], \bm X[\mathcal{N}_L(v_i)]),
\end{equation}
\begin{equation}
\mathcal{N}_L(v_i)=\{v:d(v,v_i)\le L\}
\end{equation}
where $L$ denotes the number of layers in the node-level encoder $f_i$, $d$ denotes the shortest distance between two nodes and $(\bm A[\bm\cdot;\bm\cdot], \bm X[\bm\cdot])$ selects a sub-graph from $(\bm A, \bm X)$.

\textbf{Random walk sampling} is proposed in \textbf{GCC}~\cite{qiu2020gcc} to sample sub-graphs based on random walks starting from a given node. The subset of nodes $S\in V$ is collected iteratively. At each iteration, the walk has a probability $p_{ij}$ to travel from node $v_i$ to node $v_j$ and has a probability of $p_r = 0.8$ to return to the start node. GCC considers the random walk sampling with restart as a further transformation of the $r$-ego-net centered at the start node. Given the center (start) node, the random walk sampling can be hence considered as a stronger sampling-based transformation than the ego-nets sampling.

\revision{\textbf{Network Schema} and \textbf{Meta-path} views are proposed in HeCo~\cite{xiao2021self} as two specific views for the contrastive learning of heterogeneous graphs. Given a target node of type $t$, the network schema view is a special case of 1-ego-nets consisting of neighbor nodes whose types are connected to the target node type in the network schema, and excluding nodes with the same type $t$. When computing the representation for a network schema view, aggregations are computed individually for each node type. The meta-path view consists of all meta-paths between the target node and other nodes of the same type. When computing the representation for a meta-path view, nodes of other types are masked and the information is aggregated along individual meta-paths.}

\subsubsection{Discussions of Graph View Generation}

Currently, there is no theoretical analysis guiding the generation of the view for graphs. However, \citet{tian2020makes} theoretically and empirically analyze the problem in a general view and image domain, considering the generation of the view from the aspect of mutual information. In particular, a good view generation should minimize the MI between two views $I(v_1, v_2)$, subject to $I(v_1, y)=I(v_2, y)=I(x, y)$. Intuitively, to guarantee that contrastive learning works, the generated views $v_i$ should not affect the information that determines the prediction for the downstream task, under which restriction, stronger disagreement between views leads to better learning results. \revision{Following the above idea, AD-GCL~\cite{suresh2021adversial} proposes to generate views of graphs that achieve the above minimum under constraints by parameterizing the above types of transformations and propose learnable transformations. In particular, the transformations are learned in an adversarial manner - the transformation (views generator) is trained to minimize $I(v_1, v_2)$ subject to $I(v_1, y)=I(v_2, y)=I(x, y)$, whereas the encoder is trained to maximize $I(v_1, v_2)$. Following a similar principle, InfoGCL~\cite{xu2021infogcl} proposes to discretely select optimal views from a list of candidate views based on the mutual information with downstream tasks.}

From the manifold point of view, a recent analytic study~\cite{wei2020theoretical} proposes the expansion assumption and explains the data augmentation as to prompt the continuity in the neighborhood for each instance. It indicates similar requirements for the view generating by augmentation, \textit{i.e.}, \revision{an ideal augmentation should satisfy the following two conditions, 1) the (augmentation) neighborhood of an instance does not intersect the neighborhood of instances that belong to the other class in the downstream task, 2) the neighborhood of an instance should be as large as possible, subject to 1.}

To this end, the learning on datasets with different downstream tasks may benefit from different types of transformations. For example, the property of a social network to be predicted in a downstream task may be more tolerant of minor changes in node attributes, for which the feature transformations can be more suitable. On the other hand, the property of a molecule usually depends on bonds in some functional groups, for which the edge perturbation may harm the learning while the sub-graph sampling could help. Empirically, \citet{you2020graph} observes similar results. For example, edge perturbation is found contributory to the performance on social network datasets but harmful to some molecule data.

\begin{figure*}
    \centering
    \includegraphics[width=0.98\textwidth]{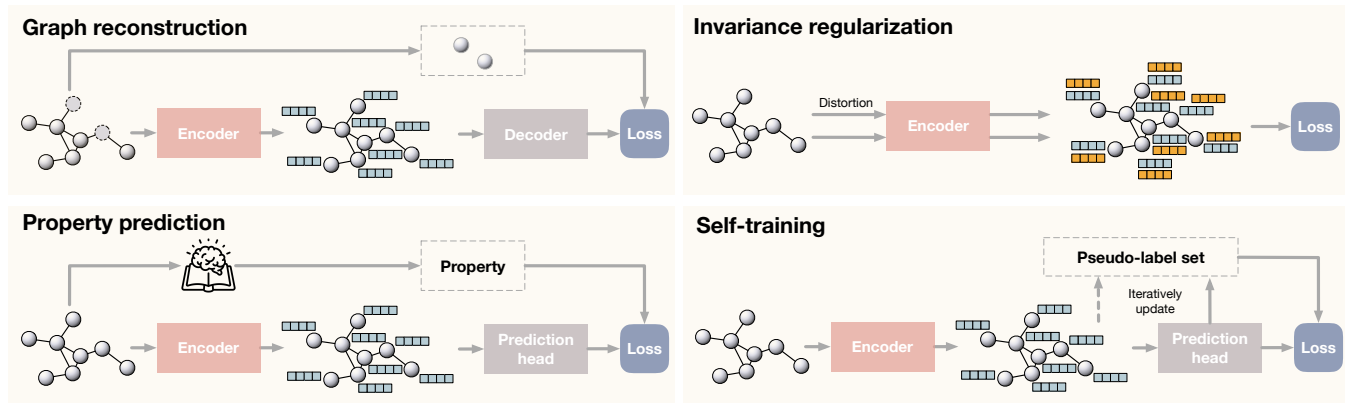}
    \caption{Illustrations of three predictive learning frameworks. For predictive learning methods, self-generated labels provide self-supervision to train the encoder together with prediction heads (or the decoder). We conclude predictive learning methods into three categories by how the prediction targets are obtained. \textbf{Top left}: the prediction targets in graph reconstruction are certain parts of given graphs. For example, GAE~\cite{kipf2016variational} performs reconstruction on the adjacency matrix, and MGAE~\cite{wang2017mgae} performs reconstruction on randomly corrupted node attributes. \textbf{Top right}: the supervision comes from the invariance regularization and additional constraints that are derived based on different theoretical frameworks and promote the learning of informative representations. \textbf{Bottom left}: the prediction targets in property prediction models are implicit and informative properties of given graphs. For example, S$^2$GRL~\cite{peng2020self} predicts k-hop connectivity between two given nodes. Moreover, GROVER~\cite{rong2020grover} utilizes motifs (functional groups) of molecules based on domain-knowledge as prediction targets. \textbf{Bottom right}: the prediction targets in self-training are pseudo-labels. In M3S~\cite{sun2020multi}, the graph neural network is trained iteratively on a pseudo-label set initialized as the set of given ground-truth labels. Clustering and prediction are conducted to update the pseudo-label set based on which a fresh graph neural network is then trained. Such operations are performed multiple times as multi-stage.}
    \label{fig:predictive}
\end{figure*}

\section{Predictive Learning}
Compared with contrastive learning methods, predictive learning methods train the graph encoder $f$ together with a prediction head $g$ under the supervision of informative labels self-generated for free. We use the term ``predictive'' instead of ``generative'' categorized by~\citet{liu2020self} to avoid confusion, as not all methods introduced in this section are necessarily generative models. Categorized by how the prediction targets are obtained, we summarize predictive learning frameworks for graphs into (1) graph reconstruction that learns to reconstruct certain parts of given graphs, (2) invariance regularization that aims to directly learn robust and informative representations, (3) graph property prediction that learns to model non-trivial properties of given graphs, and (4) multi-stage self-training with pseudo-labels. In this section, we let $\bm H\in \mathbb{R}^{|V|\times q}$ denote the desired node-level representation and ${\bm h_i}$ denote the representation of node $v_i$. The general frameworks of three types of predictive learning methods are shown in Figure~\ref{fig:predictive}. We summary all predictive methods being reviewed by this survey in Supplementary Table~2 for a more clear comparison.

\subsection{Graph Reconstruction}
Graph reconstruction provides a natural self-supervision for the training of graph neural networks. The prediction targets in graph reconstruction are certain parts of the given graphs such as the attribute of a subset of nodes or the existence of edge between a pair of nodes. In graph reconstruction tasks, the prediction head $g$ is usually called the decoder which reconstructs a graph from its representation.

\subsubsection{Non-Probabilistic Graph Autoencoders}
The autoencoders, firstly proposed in~\cite{kramer1991nonlinear}, have been widely applied for the learning of data representations. Given the success in the image domain and natural language modeling, various variations of graph autoencoder~\cite{book:hamilton} are proposed to learn graph representations. Aiming at learning the graph encoder $f$, graph autoencoders are trained to reconstruct certain parts of an input graph, given restricted access to the graph or under certain regularization to avoid identical mapping.

\textbf{GAE}~\cite{kipf2016variational} represents the simplest version of the graph autoencoders. It performs the reconstruction on the adjacency matrix $\bm A$ from the input graph $(\bm A, \bm X)$. Formally, it computes the reconstructed adjacency matrix $\hat{\bm A}$ by
\begin{align}
    \hat{\bm A} &= g(\bm H) = \sigma(\bm H\bm H^T),\\
    \bm H &= f(\bm A, \bm X),
\end{align}
and is optimized by the binary cross-entropy loss between $\hat{\bm A}$ and $\bm A$. As GAE is originally proposed to learn node-level representations for link prediction problem, it assumes two linked nodes should have similar representations. \textbf{GraphSAGE}~\cite{hamilton2017inductive} introduces a similar framework with the self-supervision of the adjacency matrix based on a different objective including negative sampling. In addition, a recent work \textbf{SuperGAT}~\cite{kim2021how} includes the GAE objective as a self-supervised auxiliary loss during training a graph attention network to guide the learning of more expressive attention operators. Similarly, \textbf{SimP-GCN}~\cite{jin2020node} applies node-pair similarity as a substitute of the adjacency matrix to construct a self-supervised auxiliary task.

\textbf{MGAE}~\cite{wang2017mgae} follows the idea of \textbf{denoising autoencoder}~\cite{vincent2010stacked}. Given a graph $(\bm A, \bm X)$, MGAE performs reconstructions on multiple randomly corrupted feature matrices $\{\tilde{\bm X}_i\}_{i=1}^m$ with a single-layer autoencoder $f_{\theta}$ and the objective
\begin{equation}
    \sum_{i=1}^m\norm{\bm X-f_{\theta}(\bm A, \tilde{\bm X}_i)}^2 + \lambda\norm{\theta}^2,
\end{equation}
where $\theta$ denotes the weights in the single-layer encoder, $\lambda$ denotes the hyper-parameter for $l^2$-regularization, and $\bm H_i := f_{\theta}(\bm A, \tilde{\bm X}_i)$ is considered as the reconstructed representation. To enable non-linearity, \cite{wang2017mgae} proposes to stack multiple single-layer autoencoders. Formally, given the reconstructed representation $\bm H^{(\ell-1)}$ at the $(\ell-1)$-th layer, the $\ell$-th layer is trained by optimizing
\begin{align}
    \sum_{i=1}^m&\norm{\bm H^{(\ell-1)} -\bm H^{(\ell)}_i}^2 + \lambda\norm{\theta_\ell}^2,\\
    &\bm H^{(\ell)}_i = f_{\theta_\ell}\left(\bm A, \tilde{\bm H}^{(\ell-1)}_i\right),
\end{align}
where $\tilde{\bm h}^{(\ell-1)}_i$ denotes the corrupted representation from the $(\ell-1)$-th layer. The reconstructed representation at the last layer is then considered as the representation for downstream tasks.

\textbf{GALA}~\cite{park2019symmetric} introduces a multi-layer autoencoder with symmetric encoder and decoder, unlike GAE and MGAE. Motivated by the Laplacian smoothing~\cite{taubin1995signal} effect of GCN encoders, GALA designs the decoder by performing Laplacian sharpening~\cite{taubin1995signal}, which prompts the decoded representation of each node to be dissimilar to the centroid of its neighbors. A Laplacian sharpening layer in the decoder $g$ in computed by
\begin{equation}
    \hat{\bm X}^{(\ell)} = 2\hat{\bm X}^{(\ell-1)} - \bm D^{-1}\bm A\hat{\bm X}^{(\ell-1)},
\end{equation}
where $\hat{\bm X}^{(\ell)}$ and $\hat{\bm X}^{(\ell-1)}$ denote the decoded representation and $D$ denotes the degree matrix. GALA reconstructs the feature matrix by optimizing the mean squared error $\norm{\hat{\bm X}-\bm X}^2$ with
\begin{align}
    \hat{\bm X} = g(\bm A, \bm H),\quad \bm H = f(\bm A, \bm X).
\end{align}

\textbf{Attribute masking}~\cite{Hu2020Strategies}, also referred to as \textbf{graph completion}~\cite{you2020does}, is another strategy to pre-train graph encoder $f$ under the graph autoencoder framework by reconstructing masked node attributes. The encoder $f$ computes the node-level representations $\bm H$ given the graph with its node attributes randomly masked. And a linear projection is applied to $\bm H$ as the decoder $g$ to reconstruct the masked attributes. When the edge attributes are also available, one can also perform reconstruction on the masked edge attributes. Although the attribute masking is not explicitly named as graph autoencoders, we categorize it as graph autoencoders since the encoders are trained by performing reconstruction on the entire or certain parts of the input graph.

\subsubsection{Variational Graph Autoencoders}
Although sharing a similar encoder-decoder structure with standard autoencoders, variational autoencoders as generative models are built upon a different mathematical foundation assuming an existing prior distribution of latent representation that generates the observed data. Primarily targeted in learning the generation of the observed data, variational graph autoencoders also shown promising performance in learning good graph representations.

\textbf{VGAE}~\cite{kipf2016variational} introduces the simplest version of variational graph autoencoders. VGAE performs reconstruction on the adjacency matrix and is composed a inference model (encoder) $q(\bm H|\bm A, \bm X)=\prod_{i=1}^{|V|} \mathcal{N}(\bm h_i|\bm\mu_i(\bm A, \bm X), \bm\Sigma_i(\bm A, \bm X))$ parameterized by graph neural networks $\bm\mu, \bm\Sigma$ and a generative model (decoder) $p(\bm A|\bm H)$ modeled by the inner product of latent variables. VGAE optimizes the variational lower bound
\begin{equation}
    \mathbb{E}_{q(\bm H|\bm A, \bm X)}\left[\log p(\bm A|\bm H)\right] - \mbox{KL}\left[q(\bm H|\bm A, \bm X)||p(\bm H)\right],
\end{equation}
where $\mbox{KL}(\cdot)$ denotes the KL-divergence and $p(\bm H)$ denotes the prior given by the Gaussian distribution.

\textbf{ARGA/ARVGA}~\cite{pan2018adversarially} propose to regularize the autoencoder with an adversarial network~\cite{goodfellow2014generative} which enforces the distribution of the latent variable to match the Gaussian prior. In addition to the encoder and decoder, a discriminator is trained to distinguish fake data generated by the encoder and the real data sampled from the Gaussian distribution. As the adversarial regularization is provably an equivalence of the JS-divergence between the distribution of the latent variable and the Gaussian prior, ARGA/ARVGA can achieve a similar effect to VGAE but with stronger regularization.

\textbf{SIG-VAE}~\cite{hasanzadeh2019semi} replace the inference model in the variational graph autoencoder a hierarchy of multiple stochastic layers to enable more flexible model of the latent variable. In particular, the inference model is given by $p(\bm H|\bm A, \bm X)=p(\bm H|\bm A, \bm X, \bm\mu, \bm\Sigma)$, where $\bm\mu$ and $\bm\Sigma$ are considered as random variables computed by stacked stochastic layers with noise injected to each layer. The marginalized $p(\bm H|\bm A, \bm X)$ is hence not necessarily a Gaussian distribution and enabled higher flexibility and expressivity.

There exist other variations of the variational graph autoencoders such as JTVAE~\cite{jin2018junction} and DGVAE~\cite{li2020dirichlet}. However, those variational graph autoencoders focus on the generation of graphs. As we mainly consider the learning of representations, we omit the introduction to those methods.

\subsubsection{Autoregressive Reconstruction}
Following the idea of \textbf{GPT}~\cite{radford2019language} for the generative pre-training of language models, \textbf{GPT-GNN}~\cite{hu2020gpt} proposes a autoregressive framework to perform reconstruction on given graphs. Ash both variational autoencoders and the autoregressive models are generative and based on reconstruction, graph autoregressive models differ from that they perform reconstruction iteratively. In particular, GPT-GNN consists of a graph encoder $f$, decoders $g_n$ and $ g_e$ for node generation and edge generation, respectively. Given a graph with its nodes and edges randomly masked, GPT-GNN generates one masked node and its edges at a time and optimizes the likelihood of the node and edges generated in the current iteration. GPT-GNN iteratively generates nodes and edges until all masked nodes are generated.

\subsection{Representation Invariance Regularization}
\revision{Adopting predictive objectives based on invariance regularization is recently trending for both image and graph domains. Methods adopting invariance regularization directly computes losses on representations and usually follows a similar framework to contrastive learning, \textit{i.e.}, to obtain two augmented graphs of the given graph, and compute the representations of the two graphs, but their objective does not include any contrast \textbf{nor requires paired or negative samples}. Hence they are categorized as predictive approaches. In particular, the objective seeks to minimize the difference between representation of two distorted graphs, encouraging representations of the graphs to be invariant to random distortions. Certain approaches are introduced to enable the learning informative representations, preventing trivial solutions to be learned.}

\revision{Inspired by \textbf{BYOL}~\cite{grill2020bootstrap} in the image domain, \textbf{BGRL}~\cite{thakoor2021bootstrapped} proposes a variation of contrastive learning framework, which eliminates the need of negative samples. Given a mini-batch of graphs $\bm B$, it compute node representations $\bm H_{\bm x, a}$ and $\bm H_{\bm x, b}$ of two augmented graphs from each $\bm x$ in $\bm B$ and minimize the following invariance-based loss with a parametric predictor $p_\theta$}
\begin{equation}
    \mathcal{L}_{\mathrm{BYOL}} = \mathbb{E}_{\bm B\sim\mathcal{P}^N}\left[ - \frac{2}{N}\sum_{\bm x\in\bm B}\frac{[p_\theta(\bm H_{\bm x, a})]^T\bm H_{\bm x, b}}{\norm{p_\theta(\bm H_{\bm x, a})}\norm{\bm H_{\bm x, b}}}\right].
\end{equation}
\revision{As no negative sample is included, certain mechanisms and restrictions, such as updating an offline encoder with exponential moving average~\cite{grill2020bootstrap} and batch normalization, are required in the framework to prevent degenerate solutions and achieve similar effect of optimizing MI bound objectives. BGRL and BYOL are commonly acknowledged as variations of contrastive methods. While the framework of BGRL follows the typical contrastive framework, the computation of the above invariance-based objective does not require paired samples or negative samples. }

\revision{\textbf{CCA-SSG}~\cite{zhang2021canonical} proposes an invariance-based objective inspired by a well-studied idea of canonical correlation analysis~\cite{hotelling1992relations, hardoon2004canonical}. The proposed objective consists of an invariance term minimizing the difference between two representations and a decorrelation term minimizing the correlation among dimensions of the representations, fomulated as}
\begin{equation}
\begin{split}
    \mathcal{L}_{\mathrm{CCA}} &= \mathbb{E}_{\bm B\sim\mathcal{P}^N}\bigg[\norm{\bm H_a - \bm H_b}^2\\ &+ \lambda\left(\norm{\bm H_a^T\bm H_a-\bm I}^2+ \norm{\bm H_b^T\bm H_b-\bm I}^2\right)\bigg],
\end{split}
\end{equation}
\revision{where $\bm H_a$ and $\bm H_b$ are batched node representations of graphs with two augmentations $a$ and $b$. Similarly to Barlow-Twins~\cite{zbontar2021barlow}, CCA-SSG uses batched representations to estimate the correlations among different dimensions. Both CCA-SSG and Barlow-Twins encourage the learning of imformative representation by reducing the correlation or redundancies among dimensions.}

\revision{\textbf{LaGraph}~\cite{xie2022self} proposes another invariance-based objective based on the assumption that all observed graph data have their latent counterparts, analogically to inaccessible clean counterparts of observed noisy data such as images. The proposed objective is derived as an self-supervised upper bound to the supervised latent graph prediction loss, formulated as}
\begin{equation}
\begin{split}
    \mathcal{L}_{\mathrm{LaGraph}} = \mathbb{E}_{(\bm A, \bm X)\sim\mathcal{P}}&\mathbb{E}_J\bigg[\norm{\mathcal{D}(\bm A, \bm H)-\bm X}^2/|V|\\
    +\lambda\Big[&\norm{\mathds{1}_J\odot\big(\bm H - \bm H'\big)}^2/|J|\Big]^{1/2}\bigg],
\end{split}
\end{equation}
\revision{where $\mathcal{D}$ is a decoder network, $J$ is a random subset of node indices, $\bm H$ is the node representation matrix of the given graph, $\bm H'$ is the representation of the graph whose nodes in $J$ are masked, and $\mathds{1}_J\odot$ means the invariance is computed on the masked nodes only. Different from the above two methods, LaGraph computes the representations for the original graph and its masked version, instead of two randomly augmented graphs, and only computes the invariance on masked nodes. The characteristics come from the derivation of the objective.}

\revision{One intuition behind the invariance regularization-based methods is that the learned representation are expected to contain enough information of the given data but be invariance to distortions on the data. Both CCA-SSG and LaGraph discuss the relationships between invariance-based method and the Information Bottleneck principle~\cite{tishby99information} indicating the above intuition. Moreover, LaGraph further discusses its relationship with denoising autoencoder and mutual information-based methods.}

% Therefore, one can also consider BGRL as a predictive method. \cite{anonymous2022selfsupervised} provides a concrete discussion on the invariance objective of BGRL as a predictive method. 
\subsection{Graph Property Prediction}
In addition to the reconstruction, an efficient way to perform self-supervised predictive learning is to design the prediction tasks based on informative graph properties that are not explicitly provided in the graph data. Commonly applied properties for self-supervised training include topology properties, statistical properties, and domain-knowledge involved properties.

\textbf{S$^2$GRL}~\cite{peng2020self} generalizes the adjacency matrix reconstruction task to a \textbf{$k$-hop connectivity prediction} task between two given nodes, motivated by that the interaction between two nodes is not limited to their direct connection. In particular, given encoded representations of any pair of nodes, the prediction head performs classification on the absolute difference between the representations. S$^2$GRL then trains the encoder and prediction head to classify the hop counts between the pair of nodes.

\textbf{Meta-path prediction}~\cite{hwang2020self} provides a self-supervision for heterogeneous graphs, such as molecules, which include multiple types of nodes and edges. A meta-path of length $\ell$ is defined as a sequence $(t_1,\cdots, t_\ell)$ where $t_i$ denotes the type of the $i$-th edge in the path. Given two nodes in a heterogeneous graph and $K$ meta-paths, the encoder $f$ and prediction heads $g_i$ ($i=1,\cdots,K$) are trained to predict if the two nodes are connected by each of the meta-paths as a binary classification task. In~\cite{hwang2020self}, the predictions of the $K$ meta-paths are included as $K$ auxiliary learning tasks in addition to the main learning task.

\textbf{GROVER}~\cite{rong2020grover} performs self-supervised learning on molecular graph data with two predictive learning tasks. In \textbf{contextual property prediction}, the encoder and prediction head is trained to predict the ``atom-bond-count'' relationship within the $k$-hop neighborhood of a given node (atom), \emph{e.g.} ``O-double\_bond-2'' if there are two atoms ``O'' connected to the given atom with double bonds. In addition, a \textbf{graph-level motif prediction} task is applied to involve the self-supervision of domain knowledge. For molecular graphs, the motifs are instantiated by the functional groups in molecules. Given a list of motifs, the graph-level prediction head predicts the existence of each motif, as a multi-label classification task.

\subsection{Self-Training with Pseudo-Labels}
Instead of the labels obtained from input graphs, the prediction targets in self-training methods are pseudo-labels obtained from the prediction in a previous stage \revision{utilizing a small portion of labeled data~\cite{yang2020rethinking}} or even randomly initialized. The self-trained graph neural networks can be either applied under a semi-supervised setting or further fine-tuned for downstream tasks. We consider the node-level classification for an instance.

Under the node-level semi-supervised setting, the \textbf{multi-stage self-training}~\cite{li2018deeper} is proposed to utilize the labeled nodes to guide the training on unlabeled nodes. Concretely, given both the labeled node set and unlabeled node set, the graph neural network is first trained on the labeled set. After the training, it performs prediction on the unlabeled set and the predicted labels with high confidence are considered as the pseudo-labels and moved to the labeled node set. Then a fresh graph neural network is trained on the updated labeled set and the above operations are performed multiple times.

\textbf{M3S}~\cite{sun2020multi} applies DeepCluster~\cite{caron2018deep} and an aligning mechanism to generate pseudo-labels on the basis of multi-stage self-training. In particular, a $K$-mean cluster is performed on node-level representations at each stage and the labels obtained from clustering are then aligned with the given true labels. A node with clustered pseudo-label is added to the labeled set for self-training in the next stage only if it matches the prediction of the classifier in the current stage. Compared to the basic multi-stage self-training, M3S considers the DeepCluster and the aligning mechanism as a self-checking mechanism and hence provides stronger self-supervision.

\revision{\textbf{ICF-GCN}~\cite{hu2021rectifying} proposes to optimize the GCN model and pseudo-labels for nodes simultaneously in an Expectation-Maximization (EM) manner. In particular, the E-step updates the GCN based on the given pseudo-labels whereas the M-step updates the pseudo-labels based on the GCN predictions. Similarly to M3S, ICF-GCN performs clustering on hidden representations to obtain GCN predicted classes. To avoid the alignment issue, both pseudo-labels and the clustered node classes are represented in relational matrix of shape $|V|\times |V|$, where an element value 1 indicates two nodes belong to the same class and 0 indicates different classes.}

A recent study~\cite{wei2020theoretical} provides the theoretical justification for the self-training with pseudo-labels based on an assumption of the expansion property and generalizes the self-training methods from semi-supervised settings into the unsupervised setting. Intuitively, the examples with correct pseudo-labels will be utilized to denoise the incorrectly pseudo-labeled examples and high accuracy can be achieved due to the expansion assumption. Under the unsupervised setting, it theoretically shows that a classifier trained with arbitrarily assigned pseudo-labels can still achieve good accuracy for some permutation of the classes.

\section{Summary of Learning Tasks and Datasets}
The self-supervised learning methods are usually applied to and evaluated on two common types of graph-related learning tasks, the graph-level inductive learning and the node-level transductive learning. The graph-level inductive learning aims to learn models predicting graph-level properties, and the models are trained and perform prediction on different sets of graphs. On the other hand, the node-level transductive learning aims to learn models performs node-level property prediction, trained and performing prediction on the same sets of large graphs. In this section, we summarize datasets under the two types of learning tasks. The statistics of common datasets are shown in Table~\ref{tab:datasets}.

\begin{table*}[th]\label{tab:datasets}
\caption{Summary and statistics of common graph datasets for self-supervised learning. Unsupervised classification refers to performing unsupervised representation learning followed by linear classification.}
\centering
\begin{tabular}{llcccccc}
\toprule
Datasets & Learning tasks  & Task level                                                                                                                         & Category                  & \# graphs & Avg. nodes & Avg. edges & \# classes     \\ \hline
\textbf{NCI1}     & \multirow{9}{*}{\begin{tabular}[c]{@{}l@{}}Unsupervised or\\ semi-supervised\\classification\end{tabular}}  & \multirow{9}{*}{Graph}                 & Small molecules           & 4110      & 29.87      & 32.30      & 2              \\
\textbf{MUTAG}    &                                                                                                      &                                        & Small molecules           & 1113      & 17.93      & 19.79      & 2              \\
\textbf{PTC-MR}   &                                                                                                      &                                        & Small molecules & 344     & 14.29      & 14.69      & 2              \\
\textbf{PROTEINS} &                                                                                                      &                                        & Bioinformatics (proteins) & 1178      & 39.06      & 72.82      & 2              \\
\textbf{DD}       &                                                                                                      &                                        & Bioinformatics (proteins) & 188       & 284.32     & 715.66     & 2              \\ 
\textbf{COLLAB}   &                                                                                                      &                                        & Social networks           & 5000      & 74.49      & 2457.78    & 2              \\
\textbf{RDT-B}    &                                                                                                      &                                        & Social networks           & 2000      & 429.63     & 497.75     & 2              \\
\textbf{RDT-M5K}  &                                                                                                      &                                        & Social networks           & 4999      & 508.52     & 594.87     & 5              \\
\textbf{IMDB-B}   &                                                                                                      &                                        & Social networks           & 1000      & 19.77      & 96.53      & 2              \\ \hline
\textbf{BBBP}     & \multirow{8}{*}{\begin{tabular}[c]{@{}l@{}}Unsupervised \\ transfer learning \\ for classification \end{tabular}}           & \multirow{8}{*}{Graph}                 & Small molecules           & 2039      & 24.05      & 25.94      & 2              \\
\textbf{Tox21}    &                                                                                                      &                                        & Small molecules           & 7831      & 18.51      & 25.94      & 12 (multi-label)  \\
\textbf{ToxCast}  &                                                                                                      &                                        & Small molecules           & 8575      & 18.78      & 19.26      & 167 (multi-label) \\
\textbf{SIDER}    &                                                                                                      &                                        & Small molecules           & 1427      & 33.64      & 35.36      & 27 (multi-label)  \\
\textbf{ClinTox}  &                                                                                                      &                                        & Small molecules           & 1478      & 26.13      & 27.86      & 2              \\
\textbf{MUV}      &                                                                                                      &                                        & Small molecules           & 93087     & 24.23      & 26.28      & 17 (multi-label)  \\
\textbf{HIV}      &                                                                                                      &                                        & Small molecules           & 41127     & 25.53      & 27.48      & 2              \\
\textbf{BACE}     &                                                                                                      &                                        & Small molecules           & 1513      & 34.12      & 36.89      & 2              \\ \hline
\textbf{CORA}     & \multirow{7}{*}{\begin{tabular}[c]{@{}l@{}}Unsupervised or\\ semi-supervised \\ classification \\ (transductive)\end{tabular}}  & \multirow{3}{*}{Node/Link}                 & Citation network          & 1         & 2,708      & 5,429      & 7              \\
\textbf{CITESEER} &                                                                                                      &                                        & Citation network          & 1         & 3,327      & 4,732      & 6              \\
\textbf{PUBMED}   &                                                                                                      &                                        & Citation network          & 1         & 19,717     & 44,338     & 3              \\ \cline{3-8}
\textbf{Coauthor CS}   &                                                                                                 &\multirow{4}{*}{Node}                   & Citation network          & 1         & 18,333     & 81,894     & 15             \\ 
\textbf{Coauthor Phy.}   &                                                                                                      &                                 & Citation network          & 1         & 34,493     & 247,962     & 5              \\ 
\textbf{Amazon Photos}   &                                                                                                      &                                 & E-commerce network          & 1         & 7,650     & 119,081     & 8              \\ 
\textbf{Amazon Comp.}   &                                                                                                      &                                  & E-commerce network          & 1         & 13,752     & 245,861    & 10             \\ \hline
\textbf{PPI}      & \multirow{3}{*}{\begin{tabular}[c]{@{}l@{}}Unsupervised \\ classification\\ (inductive)\end{tabular}}  & \multirow{3}{*}{Node}                                 & Bioinformatics (proteins) & 24        & 56,944     & 818,716    & 121 (multi-label) \\
\textbf{Flickr}   &                                                                                                      &                                        & Social network          & 1         & 89,250     & 899,765  & 7              \\ 
\textbf{Reddit}   &                                                                                                      &                                        & Social network          & 1         & 232,965    & 11,606,919   & 50              \\ 
\bottomrule
\end{tabular}
\end{table*}

\subsection{Graph-Level Inductive Learning}
Graph-level learning tasks are performed as inductive learning tasks on multiple graphs~\cite{velikovi2019deep}. Commonly used datasets for graph-level learning tasks can be divided into three types, chemical molecule datasets, protein datasets, and social network datasets.

\textbf{Chemical Molecular Property Prediction}. In a molecular graph, each node represents an atom in a molecule where the atom index is indicated by the node attribute and each edge represents a bond in the molecule. Datasets for chemical molecular property prediction are also categorized as small molecule datasets in TUDataset~\cite{Morris2020tudataset}. Traditional molecule classification datasets such as \textbf{NCI1}~\cite{NCI1} and \textbf{MUTAG}~\cite{mutag} are the most commonly used datasets for unsupervised graph representation learning in self-supervision related studies~\cite{you2020graph, hassani2020contrastive}. In addition, the molecule property prediction models can be also trained in a self-supervised pre-training and finetuning fashion for semi-supervised learning and transfer learning. Recent works~\cite{Hu2020Strategies, you2020graph, rong2020grover} build their pre-training molecule dataset by sampling unlabeled molecules from the \textbf{ZINC15}~\cite{sterling2015zinc} database containing 750 million molecules. \textbf{MoleculeNet}~\cite{wu2018moleculenet} also provides a collection of molecular graph datasets for molecule property prediction, which is suitable for downstream graph classification. Among all the datasets in MoleculeNet, the classification datasets such as BBBP, Tox21, and HIV are used for the evaluation of several self-supervised learning methods~\cite{Hu2020Strategies, you2020graph, rong2020grover}.

\textbf{Protein Biological Function Prediction}. The protein is a particular type of molecule but is represented differently by graph data. In a protein graph, nodes represent amino acids and an edge indicates the two connected nodes are less than 6 Angstroms apart. Datasets for chemical molecular property prediction are also categorized as bioinfomatics datasets in TUDataset. Similar to the chemical molecule datasets, protein datasets can be used in both unsupervised representation learning, such as \textbf{PROTEINS}~\cite{proteins} and \textbf{DD}~\cite{DD}, and in the two-stage training. 

\textbf{Social Network Property Prediction}. A social network graph dataset considers each entity (\emph{e.g.} a user or an author) as a node and their social connections as edges. As social networks in different datasets represent differently, social network graph datasets are not typically used for transfer learning. Social network graph datasets used in recent self-supervised studies~\cite{qiu2020gcc, you2020graph} are typical datasets for graph classification~\cite{deep2015yanardag} such as \textbf{COLLAB}, \textbf{REDDIT-B} and \textbf{IMDB-B}.

\subsection{Node-Level Transductive Learning}
Node-level learning tasks can be performed as transductive learning tasks on large graphs~\cite{velikovi2019deep}, where are nodes and the complete graph structure, together with labels of a portion of nodes, are available for training. The citation network datasets~\cite{yang2016revisiting}, including \textbf{CORA}~\cite{mccallum2000cora}, \textbf{CITESEER}~\cite{citeseer} and \textbf{PUBMED}~\cite{sen2008pubmed} are commonly used for node-level transductive learning. There are three typical ways to use the citation network datasets. Contrastive learning methods~\cite{velikovi2019deep, sun2019infograph, hassani2020contrastive} are usually evaluated on the social network datasets by performing unsupervised representation learning followed by a linear classification with fixed representations, whereas predictive learning studies usually perform unsupervised representation learning followed by clustering~\cite{wang2017mgae, park2019symmetric} or semi-supervised link prediction~\cite{kipf2016variational, peng2020self} on the social network datasets.

\revision{Motivated by the concern that current GNN evaluations becomes saturated on above citation network datasets, \citet{shchur2018pitfalls} construct four additional node-level datasets, \textbf{Coauthor-CS}, \textbf{Coauthor-Physics} from the Microsoft Academic Graph~\cite{sinha2015overview}, \textbf{Amazon-Photos}, and \textbf{Amazon-Computers} from the Amazon Co-purchase Graph~\cite{mcauley2015image}. The four datasets contain larger graphs with more nodes and edges. And their learning tasks are hence more challenging compared to the citation network datasets. }

\subsection{Node-Level Inductive Learning}
\revision{Node-level inductive learning performs training and testing on separate subsets. There are two typical ways to split the nodes for training and testing. First, in cases that all nodes are from the same large graph, a random subset of nodes are selected for testing and is masked out together with their edges during training, in contrast to the transductive learning where all nodes and the complete graph structure are used during training. Two commonly used node-level inductive learning datasets in the first case are \textbf{Reddit}~\cite{hamilton2017inductive} and \textbf{Flickr}~\cite{graphsaint-iclr20}. Each node in Reddit represents a post and posts are connected by edges if they are commented by the same user. The node classification task is to predict which community the post belongs to. For the Flickr dataset, each node represents an image uploaded to Flickr and an edge between nodes indicates that they share common properties such as the same geographic location or commented by the same user. The task is to predict the class (based on tags) each image belongs to.}

In cases that all nodes belong to separate graphs, the training and testing nodes are split by graphs.
\citet{zitnik2017predicting} build a inductive learning dataset in the second case containing 395K unlabeled protein obtained from \textbf{protein–protein interaction} (\textbf{PPI}) \textbf{networks} and perform finetuning on PPI networks consisting of 88K proteins labeled with 40 fine-grained biological functions, obtained from \citet{zitnik2019evolution}. Each node in a graph represents a protein and an edge indicates the existance of interaction between the proteins. The task of PPI is to predict the gene ontology sets a protein belongs to.

% \begin{figure*}[t]
%     \centering
%     \includegraphics[width=0.95\textwidth]{figures/sslgraph.pdf}
%     \caption{An overview of the developed sslgraph library within DIG: Dive into Graphs. The library provides a standardized evaluation framework consisting of customizable framework and pre-implemented models for contrastive methods, data interface and evaluation tools for both contrastive and predictive methods.}
%     \label{fig:sslgraph}
% \end{figure*}

\section{An Open-source Library}
We develop an open-source library DIG: Dive into Graphs~\cite{liu2021dig}\footnote[1]{The open-source library is now publicly available at the URL: https://github.com/divelab/DIG/}, which includes a module, known as sslgraph, for self-supervised learning of GNNs. DIG-sslgraph is based on Pytorch~\cite{Paszke2019Pytorch} and Pytorch Geometric~\cite{Fey2019Geometric} and aims at easy implementation and standardized evaluation of SSL methods for graph neural networks. In particular, we provide a unified and highly customizable framework for contrastive learning methods, standardized data interface, and evaluation tools that are compatible for evaluating both contrastive methods and predictive methods. The overview of the library is shown in Supplementary Figure~1.

Given the developed unified contrastive framework as a base class, particular contrastive learning methods can be easily implemented by specifying their objective, functions for view generation, and their encoders. We also pre-implement four state-of-the-art contrastive methods for either node-level or graph-level tasks based on the unified framework, including InfoGraph~\cite{sun2019infograph}, GRACE~\cite{zhu2020deep}, MVGRL~\cite{sun2020multi}, and GraphCL~\cite{you2020graph}. The provided data interface includes datasets from TUDataset~\cite{Morris2020tudataset} and the citation network dataset~\cite{yang2016revisiting}. Other datasets from Pytorch Geometric and new datasets processed by Pytorch Geometric are also supported by our data interface. The provided evaluation tools and data interface allow standardized evaluations of SSL methods and fair comparisons with existing SSL methods under common evaluation settings, including unsupervised graph-level representation learning, semi-supervised graph classification (or transfer learning, depending on the datasets), and unsupervised node-level representation learning. Altogether, our open-source library provides a complete and extensible framework for developing and evaluating SSL methods for GNNs. \revision{To show the efficiency and effectiveness of the DIG-sslgraph library, we compare the training time, memory consumption, and downstream accuracy of four SSL methods on multiple datasets between the original implementations and DIG counterparts. The results are shown and discussed in Appendix G.}

\section{Challenges and Future Directions}
\revision{While existing SSL approaches have shown promising effectiveness on learning from graph data, there still exist several challenges due to the more complicated structure and more diverse tasks of graphs.
In this section, we discuss the remaining challenges as well as potential directions for future studies on Graph self-supervised learning.}

\revision{\textbf{The optimal views generation w.r.t specific downstream tasks are still unclear for contrastive methods}. The performance of the learned representation or pre-trained model on downstream tasks heavily depends on the selection of transformations to generate views. The optimal view generation also depends on specific downstream tasks. However, there is still no way to obtain the optimal view for each downstream task even the task is available. Several studies have explored approaches to utilizing better views for contrast based on adversarial learning~\cite{suresh2021adversial} or searching~\cite{xu2021infogcl}, but views generated by the two approaches are still not optimal due to their limited search space for graph transformations. In particular, \citet{xu2021infogcl} only considers a finite set of graph transformations whose search space is also limited. In addition, \citet{suresh2021adversial} only considers parameterized structural transformations, and more complicated learnable view generation involving feature-space, structure-space, and sampling-based transformation are still challenging and are limited by the development of graph generation studies~\cite{you2018graphrnn, Shi2020GraphAF:, zhang2020moflow, luo2021graphdf}. Moreover, the above methods require available downstream tasks at the pretraining stage. They become inapplicable when downstream tasks are unavailable or in the unsupervised representation learning setting. Hence it is also desired to study universally optimal views under the downstream task-agnostic setting.}

\revision{\textbf{There is no unified theory or theoretical framework for predictive methods}. Unlike contrastive methods grounded by the problem of mutual information maximization, the predictive methods, especially the graph property prediction and invariance regularization-based methods, utilize different pretext learning tasks motivated by individual hypotheses and based on empirical studies. However, they lack guidance from unified theoretical frameworks to design specific pretext tasks for different downstream tasks. The information bottleneck principle may be used to interpret the effectiveness of several predictive methods but further study and investigation are desired.}

\revision{\textbf{Richer domain knowledge can be better utilized as self-supervision}. For graph machine learning tasks oriented by other research areas such as biomedical researches and quantum physics, existing constraints and rules from domain knowledge naturally contain rich supervision benefiting the learning of downstream tasks. Including domain knowledge has shown to be effective for both contrastive methods~\cite{li2021pairwise} and predictive methods~\cite{rong2020grover}. Currently, only simple domain knowledge-based tasks such as the motif prediction~\cite{rong2020grover} are adopted. Future studies on designing novel tasks better utilizing domain knowledge such as functional groups and quantum mechanisms are promising directions.}

\revision{\textbf{Scaling-up and efficiency issues are to be addressed}. Many existing SSL approaches suffer from memory issues and computing efficiency issues in terms of time consumption when scaled up to larger graphs. For contrastive methods, the scaling-up issue becomes more critical as their performance usually relies on a larger number of samples in a mini-batch. In addition, as the contrastive framework requires computing representations of multiple views, their memory consumption is times higher than predictive approaches. When the computing of view generation for contrastive methods or the graph properties computation for predictive methods is heavy, the computation time may increase sharply as the graph scales up. The above issues prevent the application of existing methods to extremely large graphs in industrial scenarios or other research areas (\textit{e.g.}, protein networks and particles in materials). Currently, studies addressing the scaling-up issue for SSL methods are still lacking and less explored.}

\newrevision{\textbf{Explainability of SSL for GNN requires further studies}. The explainability for GNNs is critical in multiple application scenarios to assure the reliability and security of GNN models. For example, in drug discovery, it is important to understand which functional group in a molecular graph leads to the GNN decision for a property prediction. A survey work~\cite{yuan2020explainability} provides a thorough introduction to existing explanation methods for GNNs. However, existing SOTA studies focus on the explanation under supervised setting and require downstream tasks to perform explanation, and only limited methods, such as gradient-based methods, can be adapted to explain pre-trained GNN encoders without given the downstream prediction head. A recent work~\cite{xie2022task} proposes the task-agnostic explanation framework to enable high-quality GNN explanations without downstream tasks. The task-agnostic framework can be utilized to explain GNNs trained through SSL. More studies and investigations are desired in this direction.}

\section{Conclusion}
Despite recent successes in natural language processing and computer vision, the self-supervised learning applied to graph data is still an emerging field and has significant challenges to be addressed. Existing methods employ self-supervision to graph neural networks through either contrastive learning or predictive learning. We summarize current self-supervised learning methods and provide unified reviews for the two approaches. We unify existing contrastive learning methods for GNNs with a general contrastive framework, which performs mutual information maximization on different views of given graphs with appropriate MI estimators. We demonstrate that any existing method can be realized by determinating its MI estimator, views generation, and graph encoder. We further provide detailed descriptions of existing options for the components and discussions to guide the choice of those components. For predictive learning, we categorize existing methods into graph reconstruction, property prediction, and self-training based on how labels are generated from the data. A thorough review is provided for methods in all three types of predictive learning. Finally, we summarize common graph datasets of different domains and introduce what learning tasks the datasets are involved in to provide a clear view to conduct future evaluation experiments. Altogether, our unified treatment of SSL in GNNs in terms of methodologies, datasets, evaluations, and open-source software is anticipated to foster methodological developments and facilitate empirical comparisons.

% \appendices
% \section{Proof of the First Zonklar Equation}
% Appendix one text goes here.

% you can choose not to have a title for an appendix
% if you want by leaving the argument blank
% \section{}
% Appendix two text goes here.

% use section* for acknowledgment
\ifCLASSOPTIONcompsoc
  % The Computer Society usually uses the plural form
  \section*{Acknowledgments}
\else
  % regular IEEE prefers the singular form
  \section*{Acknowledgment}
\fi

This work was supported in part by National Science Foundation grants IIS-2006861 and DBI-2028361, and
National Institutes of Health grant 1R21NS102828. We thank the scientific community for providing valuable feedback and comments, which lead to improvements of this work.

% Can use something like this to put references on a page
% by themselves when using endfloat and the captionsoff option.
\ifCLASSOPTIONcaptionsoff
  \newpage
\fi

% trigger a \newpage just before the given reference
% number - used to balance the columns on the last page
% adjust value as needed - may need to be readjusted if
% the document is modified later
%\IEEEtriggeratref{8}
% The "triggered" command can be changed if desired:
%\IEEEtriggercmd{\enlargethispage{-5in}}

% references section

% can use a bibliography generated by BibTeX as a .bbl file
% BibTeX documentation can be easily obtained at:
% http://mirror.ctan.org/biblio/bibtex/contrib/doc/
% The IEEEtran BibTeX style support page is at:
% http://www.michaelshell.org/tex/ieeetran/bibtex/
%\bibliographystyle{IEEEtran}
% argument is your BibTeX string definitions and bibliography database(s)
%\bibliography{IEEEabrv,../bib/paper}
%
% <OR> manually copy in the resultant .bbl file
% set second argument of \begin to the number of references
% (used to reserve space for the reference number labels box)

\bibliographystyle{IEEEtranN}
\bibliography{reference,dive}

% Generated by IEEEtranN.bst, version: 1.14 (2015/08/26)
\begin{thebibliography}{135}
\providecommand{\natexlab}[1]{#1}
\providecommand{\url}[1]{#1}
\csname url@samestyle\endcsname
\providecommand{\newblock}{\relax}
\providecommand{\bibinfo}[2]{#2}
\providecommand{\BIBentrySTDinterwordspacing}{\spaceskip=0pt\relax}
\providecommand{\BIBentryALTinterwordstretchfactor}{4}
\providecommand{\BIBentryALTinterwordspacing}{\spaceskip=\fontdimen2\font plus
\BIBentryALTinterwordstretchfactor\fontdimen3\font minus
  \fontdimen4\font\relax}
\providecommand{\BIBforeignlanguage}[2]{{%
\expandafter\ifx\csname l@#1\endcsname\relax
\typeout{** WARNING: IEEEtranN.bst: No hyphenation pattern has been}%
\typeout{** loaded for the language `#1'. Using the pattern for}%
\typeout{** the default language instead.}%
\else
\language=\csname l@#1\endcsname
\fi
#2}}
\providecommand{\BIBdecl}{\relax}
\BIBdecl

\bibitem[Yang and Xu(2020)]{yang2020rethinking}
Y.~Yang and Z.~Xu, ``Rethinking the value of labels for improving
  class-imbalanced learning,'' in \emph{Advances in Neural Information
  Processing Systems}, 2020.

\bibitem[Ulyanov et~al.(2018)Ulyanov, Vedaldi, and Lempitsky]{Ulyanov2018deep}
D.~Ulyanov, A.~Vedaldi, and V.~Lempitsky, ``Deep image prior,'' in
  \emph{Proceedings of the IEEE Conference on Computer Vision and Pattern
  Recognition}, 2018.

\bibitem[Xie et~al.(2020)Xie, Wang, and Ji]{xie2020noise2same}
Y.~Xie, Z.~Wang, and S.~Ji, ``Noise2{S}ame: Optimizing a self-supervised bound
  for image denoising,'' in \emph{Advances in Neural Information Processing
  Systems}, vol.~33, 2020, pp. 20\,320--20\,330.

\bibitem[Laine et~al.(2019)Laine, Karras, Lehtinen, and Aila]{laine2019high}
S.~Laine, T.~Karras, J.~Lehtinen, and T.~Aila, ``High-quality self-supervised
  deep image denoising,'' \emph{Advances in Neural Information Processing
  Systems}, vol.~32, pp. 6970--6980, 2019.

\bibitem[Krull et~al.(2019)Krull, Buchholz, and Jug]{krull2019noise2void}
A.~Krull, T.-O. Buchholz, and F.~Jug, ``Noise2{V}oid-learning denoising from
  single noisy images,'' in \emph{Proceedings of the IEEE Conference on
  Computer Vision and Pattern Recognition}, 2019, pp. 2129--2137.

\bibitem[Batson and Royer(2019)]{batson2019noise2self}
J.~Batson and L.~Royer, ``Noise2{S}elf: Blind denoising by self-supervision,''
  in \emph{Proceedings of the 36th International Conference on Machine
  Learning}, vol.~97, 2019, pp. 524--533.

\bibitem[Devlin et~al.(2019)Devlin, Chang, Lee, and Toutanova]{devlin2019bert}
J.~Devlin, M.~Chang, K.~Lee, and K.~Toutanova, ``{BERT:} pre-training of deep
  bidirectional transformers for language understanding,'' in \emph{Proceedings
  of the 2019 Conference of the North American Chapter of the Association for
  Computational Linguistics: Human Language Technologies}, 2019, pp.
  4171--4186.

\bibitem[Wu et~al.(2019)Wu, Wang, and Wang]{wu2019self}
J.~Wu, X.~Wang, and W.~Y. Wang, ``Self-supervised dialogue learning,'' in
  \emph{Proceedings of the 57th Annual Meeting of the Association for
  Computational Linguistics}, 2019, pp. 3857--3867.

\bibitem[Wang et~al.(2019)Wang, Wang, Xiong, Yu, Guo, Chang, and
  Wang]{wang2019self}
H.~Wang, X.~Wang, W.~Xiong, M.~Yu, X.~Guo, S.~Chang, and W.~Y. Wang,
  ``Self-supervised learning for contextualized extractive summarization,'' in
  \emph{Proceedings of the 57th Annual Meeting of the Association for
  Computational Linguistics}, 2019, pp. 2221--2227.

\bibitem[Hassani and Khasahmadi(2020)]{hassani2020contrastive}
K.~Hassani and A.~H. Khasahmadi, ``Contrastive multi-view representation
  learning on graphs,'' in \emph{Proceedings of the 37th International
  Conference on Machine Learning}, 2020.

\bibitem[Chen et~al.(2020)Chen, Kornblith, Norouzi, and Hinton]{chen2020simple}
T.~Chen, S.~Kornblith, M.~Norouzi, and G.~Hinton, ``A simple framework for
  contrastive learning of visual representations,'' in \emph{Proceedings of the
  International Conference on Machine Learning}, 2020.

\bibitem[Oord et~al.(2018)Oord, Li, and Vinyals]{oord2018representation}
A.~v.~d. Oord, Y.~Li, and O.~Vinyals, ``Representation learning with
  contrastive predictive coding,'' \emph{arXiv preprint arXiv:1807.03748},
  2018.

\bibitem[Tian et~al.(2019)Tian, Krishnan, and Isola]{tian2019contrastive}
Y.~Tian, D.~Krishnan, and P.~Isola, ``Contrastive multiview coding,''
  \emph{arXiv preprint arXiv:1906.05849}, 2019.

\bibitem[Perozzi et~al.(2014)Perozzi, Al-Rfou, and Skiena]{Perozzi2014deepwalk}
B.~Perozzi, R.~Al-Rfou, and S.~Skiena, ``Deepwalk: Online learning of social
  representations,'' in \emph{Proceedings of the 20th ACM SIGKDD International
  Conference on Knowledge Discovery and Data Mining}, 2014, pp. 701--710.

\bibitem[Narayanan et~al.(2017)Narayanan, Chandramohan, Venkatesan, Chen, Liu,
  and Jaiswal]{narayanan2017graph2vec}
A.~Narayanan, M.~Chandramohan, R.~Venkatesan, L.~Chen, Y.~Liu, and S.~Jaiswal,
  ``graph2vec: Learning distributed representations of graphs,'' \emph{arXiv
  preprint arXiv:1707.05005}, 2017.

\bibitem[Tsitsulin et~al.(2018)Tsitsulin, Mottin, Karras, Bronstein, and
  M{\"u}ller]{tsitsulin2018sgr}
A.~Tsitsulin, D.~Mottin, P.~Karras, A.~Bronstein, and E.~M{\"u}ller, ``{SGR}:
  Self-supervised spectral graph representation learning,'' \emph{arXiv
  preprint arXiv:1811.06237}, 2018.

\bibitem[Doersch et~al.(2015)Doersch, Gupta, and
  Efros]{doersch2015unsupervised}
C.~Doersch, A.~Gupta, and A.~A. Efros, ``Unsupervised visual representation
  learning by context prediction,'' in \emph{Proceedings of the IEEE
  international conference on computer vision}, 2015, pp. 1422--1430.

\bibitem[Noroozi and Favaro(2016)]{noroozi2016unsupervised}
M.~Noroozi and P.~Favaro, ``Unsupervised learning of visual representations by
  solving jigsaw puzzles,'' in \emph{European conference on computer
  vision}.\hskip 1em plus 0.5em minus 0.4em\relax Springer, 2016, pp. 69--84.

\bibitem[He et~al.(2020)He, Fan, Wu, Xie, and Girshick]{he2020momentum}
K.~He, H.~Fan, Y.~Wu, S.~Xie, and R.~Girshick, ``Momentum contrast for
  unsupervised visual representation learning,'' in \emph{Proceedings of the
  IEEE Conference on Computer Vision and Pattern Recognition}, 2020.

\bibitem[Grill et~al.(2020)Grill, Strub, Altch\'{e}, Tallec, Richemond,
  Buchatskaya, Doersch, Avila~Pires, Guo, Gheshlaghi~Azar, Piot, kavukcuoglu,
  Munos, and Valko]{grill2020bootstrap}
J.-B. Grill, F.~Strub, F.~Altch\'{e}, C.~Tallec, P.~Richemond, E.~Buchatskaya,
  C.~Doersch, B.~Avila~Pires, Z.~Guo, M.~Gheshlaghi~Azar, B.~Piot,
  k.~kavukcuoglu, R.~Munos, and M.~Valko, ``Bootstrap your own latent - a new
  approach to self-supervised learning,'' in \emph{Advances in Neural
  Information Processing Systems}, vol.~33, 2020, pp. 21\,271--21\,284.

\bibitem[Wu et~al.(2018)Wu, Ramsundar, Feinberg, Gomes, Geniesse, Pappu,
  Leswing, and Pande]{wu2018moleculenet}
Z.~Wu, B.~Ramsundar, E.~N. Feinberg, J.~Gomes, C.~Geniesse, A.~S. Pappu,
  K.~Leswing, and V.~Pande, ``Moleculenet: a benchmark for molecular machine
  learning,'' \emph{Chemical Science}, vol.~9, no.~2, pp. 513--530, 2018.

\bibitem[Wang et~al.(2020)Wang, Liu, Luo, Xu, Xie, Wang, Cai, and
  Ji]{wang2020moleculekit}
Z.~Wang, M.~Liu, Y.~Luo, Z.~Xu, Y.~Xie, L.~Wang, L.~Cai, and S.~Ji, ``Advanced
  graph and sequence neural networks for molecular property prediction and drug
  discovery,'' \emph{arXiv preprint arXiv:2012.01981}, 2020.

\bibitem[Kipf and Welling(2017)]{kipf2017semi}
T.~N. Kipf and M.~Welling, ``Semi-supervised classification with graph
  convolutional networks,'' in \emph{International Conference on Learning
  Representations}, 2017.

\bibitem[Xu et~al.(2019)Xu, Hu, Leskovec, and Jegelka]{xu2018how}
K.~Xu, W.~Hu, J.~Leskovec, and S.~Jegelka, ``How powerful are graph neural
  networks?'' in \emph{International Conference on Learning Representations},
  2019.

\bibitem[Gao and Ji(2019{\natexlab{a}})]{Gao:ICML19}
H.~Gao and S.~Ji, ``Graph {U-Nets},'' in \emph{Proceedings of the 36th
  International Conference on Machine Learning}, 2019, pp. 2083--2092.

\bibitem[Liu et~al.(2020{\natexlab{a}})Liu, Wang, and Ji]{liu2020non}
M.~Liu, Z.~Wang, and S.~Ji, ``Non-local graph neural networks,'' \emph{arXiv
  preprint arXiv:2005.14612}, 2020.

\bibitem[Cai et~al.(2020)Cai, Li, Wang, and Ji]{cai2020line}
L.~Cai, J.~Li, J.~Wang, and S.~Ji, ``Line graph neural networks for link
  prediction,'' \emph{arXiv preprint arXiv:2010.10046}, 2020.

\bibitem[Liu et~al.(2020{\natexlab{b}})Liu, Gao, and Ji]{Liu:Deeper}
M.~Liu, H.~Gao, and S.~Ji, ``Towards deeper graph neural networks,'' in
  \emph{Proceedings of the 26th ACM SIGKDD International Conference on
  Knowledge Discovery and Data Mining}, 2020, pp. 338--348.

\bibitem[Cai and Ji(2020)]{CaiMlinkAAAI20}
L.~Cai and S.~Ji, ``A multi-scale approach for graph link prediction,'' in
  \emph{Proceedings of the 34th AAAI Conference on Artificial Intelligence},
  2020, pp. 3308--3315.

\bibitem[Gao et~al.(2018)Gao, Wang, and Ji]{Gao:KDD18}
H.~Gao, Z.~Wang, and S.~Ji, ``Large-scale learnable graph convolutional
  networks,'' in \emph{Proceedings of the 24th ACM SIGKDD International
  Conference on Knowledge Discovery and Data Mining}, 2018, pp. 1416--1424.

\bibitem[Gao et~al.(2019)Gao, Chen, and Ji]{Gao:www19}
H.~Gao, Y.~Chen, and S.~Ji, ``Learning graph pooling and hybrid convolutional
  operations for text representations,'' in \emph{Proceedings of the Web
  Conference}, 2019, pp. 2743--2749.

\bibitem[Gao et~al.(2021)Gao, Liu, and Ji]{gao2020topology}
H.~Gao, Y.~Liu, and S.~Ji, ``Topology-aware graph pooling networks,''
  \emph{IEEE Transactions on Pattern Analysis and Machine Intelligence},
  vol.~43, no.~12, pp. 4512--4518, 2021.

\bibitem[Wang and Ji(2020)]{wang2020second}
Z.~Wang and S.~Ji, ``Second-order pooling for graph neural networks,''
  \emph{IEEE Transactions on Pattern Analysis and Machine Intelligence}, 2020.

\bibitem[Yuan and Ji(2020)]{Yuan:ICLR2020}
H.~Yuan and S.~Ji, ``{StructPool}: Structured graph pooling via conditional
  random fields,'' in \emph{Proceedings of the 8th International Conference on
  Learning Representations}, 2020.

\bibitem[Gao and Ji(2019{\natexlab{b}})]{gao2019graph}
H.~Gao and S.~Ji, ``Graph representation learning via hard and channel-wise
  attention networks,'' in \emph{Proceedings of the 25th ACM SIGKDD
  International Conference on Knowledge Discovery \& Data Mining}.\hskip 1em
  plus 0.5em minus 0.4em\relax ACM, 2019, pp. 741--749.

\bibitem[Yuan et~al.(2020{\natexlab{a}})Yuan, Tang, Hu, and Ji]{Yuan:XGNN}
H.~Yuan, J.~Tang, X.~Hu, and S.~Ji, ``{XGNN}: Towards model-level explanations
  of graph neural networks,'' in \emph{Proceedings of the 26th ACM SIGKDD
  International Conference on Knowledge Discovery and Data Mining}, 2020, pp.
  430--438.

\bibitem[Liu et~al.(2020{\natexlab{c}})Liu, Yuan, Cai, and Ji]{Liu:Protein}
Y.~Liu, H.~Yuan, L.~Cai, and S.~Ji, ``Deep learning of high-order interactions
  for protein interface prediction,'' in \emph{Proceedings of the 26th ACM
  SIGKDD International Conference on Knowledge Discovery and Data Mining},
  2020, pp. 679--687.

\bibitem[Veli{\v c}kovi{\'c} et~al.(2019)Veli{\v c}kovi{\'c}, Fedus, Hamilton,
  Li{\`o}, Bengio, and Hjelm]{velikovi2019deep}
P.~Veli{\v c}kovi{\'c}, W.~Fedus, W.~L. Hamilton, P.~Li{\`o}, Y.~Bengio, and
  D.~Hjelm, ``Deep graph infomax,'' in \emph{International Conference on
  Learning Representations}, 2019.

\bibitem[Kipf and Welling(2016)]{kipf2016variational}
T.~N. Kipf and M.~Welling, ``Variational graph auto-encoders,'' \emph{arXiv
  preprint arXiv:1611.07308}, 2016.

\bibitem[Ma and Tang(2020)]{ma2020deep}
Y.~Ma and J.~Tang, \emph{Deep Learning on Graphs}.\hskip 1em plus 0.5em minus
  0.4em\relax Cambridge University Press, 2020.

\bibitem[Sun et~al.(2019{\natexlab{a}})Sun, Hoffman, Verma, and
  Tang]{sun2019infograph}
F.-Y. Sun, J.~Hoffman, V.~Verma, and J.~Tang, ``Infograph: Unsupervised and
  semi-supervised graph-level representation learning via mutual information
  maximization,'' in \emph{International Conference on Learning
  Representations}, 2019.

\bibitem[Peng et~al.(2020{\natexlab{a}})Peng, Huang, Luo, Zheng, Rong, Xu, and
  Huang]{peng2020graph}
Z.~Peng, W.~Huang, M.~Luo, Q.~Zheng, Y.~Rong, T.~Xu, and J.~Huang, ``Graph
  representation learning via graphical mutual information maximization,'' in
  \emph{Proceedings of The Web Conference 2020}, 2020, pp. 259--270.

\bibitem[Hu et~al.(2020{\natexlab{a}})Hu, Liu, Gomes, Zitnik, Liang, Pande, and
  Leskovec]{Hu2020Strategies}
W.~Hu, B.~Liu, J.~Gomes, M.~Zitnik, P.~Liang, V.~Pande, and J.~Leskovec,
  ``Strategies for pre-training graph neural networks,'' in \emph{International
  Conference on Learning Representations}, 2020.

\bibitem[Jiao et~al.(2020)Jiao, Xiong, Zhang, Zhang, Zhang, and
  Zhu]{jiao2020sub}
Y.~Jiao, Y.~Xiong, J.~Zhang, Y.~Zhang, T.~Zhang, and Y.~Zhu, ``Sub-graph
  contrast for scalable self-supervised graph representation learning,'' in
  \emph{IEEE International Conference on Data Mining}.\hskip 1em plus 0.5em
  minus 0.4em\relax IEEE, 2020, pp. 222--231.

\bibitem[Zhu et~al.(2021)Zhu, Xu, Yu, Liu, Wu, and Wang]{zhu2021graph}
Y.~Zhu, Y.~Xu, F.~Yu, Q.~Liu, S.~Wu, and L.~Wang, ``Graph contrastive learning
  with adaptive augmentation,'' in \emph{Proceedings of The Web Conference
  2021}, 2021.

\bibitem[Zhu et~al.(2020)Zhu, Xu, Yu, Liu, Wu, and Wang]{zhu2020deep}
------, ``Deep graph contrastive representation learning,'' in \emph{ICML
  Workshop on Graph Representation Learning and Beyond}, 2020.

\bibitem[Thakoor et~al.(2021)Thakoor, Tallec, Azar, Munos,
  Veli{\v{c}}kovi{\'c}, and Valko]{thakoor2021bootstrapped}
S.~Thakoor, C.~Tallec, M.~G. Azar, R.~Munos, P.~Veli{\v{c}}kovi{\'c}, and
  M.~Valko, ``Large-scale representation learning on graphs via
  bootstrapping,'' \emph{arXiv preprint arXiv:2102.06514}, 2021.

\bibitem[Qiu et~al.(2020)Qiu, Chen, Dong, Zhang, Yang, Ding, Wang, and
  Tang]{qiu2020gcc}
J.~Qiu, Q.~Chen, Y.~Dong, J.~Zhang, H.~Yang, M.~Ding, K.~Wang, and J.~Tang,
  ``{GCC}: Graph contrastive coding for graph neural network pre-training,'' in
  \emph{Proceedings of the 26th ACM SIGKDD International Conference on
  Knowledge Discovery \& Data Mining}, 2020, pp. 1150--1160.

\bibitem[You et~al.(2020{\natexlab{a}})You, Chen, Sui, Chen, Wang, and
  Shen]{you2020graph}
Y.~You, T.~Chen, Y.~Sui, T.~Chen, Z.~Wang, and Y.~Shen, ``Graph contrastive
  learning with augmentations,'' in \emph{Advances in Neural Information
  Processing Systems}, vol.~33, 2020, pp. 5812--5823.

\bibitem[Wang et~al.(2017)Wang, Pan, Long, Zhu, and Jiang]{wang2017mgae}
C.~Wang, S.~Pan, G.~Long, X.~Zhu, and J.~Jiang, ``Mgae: Marginalized graph
  autoencoder for graph clustering,'' in \emph{Proceedings of the 2017 ACM on
  Conference on Information and Knowledge Management}, 2017, pp. 889--898.

\bibitem[Park et~al.(2019)Park, Lee, Chang, Lee, and Choi]{park2019symmetric}
J.~Park, M.~Lee, H.~J. Chang, K.~Lee, and J.~Y. Choi, ``Symmetric graph
  convolutional autoencoder for unsupervised graph representation learning,''
  in \emph{Proceedings of the IEEE/CVF International Conference on Computer
  Vision}, 2019, pp. 6519--6528.

\bibitem[Pan et~al.(2018)Pan, Hu, Long, Jiang, Yao, and
  Zhang]{pan2018adversarially}
S.~Pan, R.~Hu, G.~Long, J.~Jiang, L.~Yao, and C.~Zhang, ``Adversarially
  regularized graph autoencoder for graph embedding,'' in \emph{Proceedings of
  the 27th International Joint Conference on Artificial Intelligence}, 2018,
  pp. 2609--2615.

\bibitem[Hasanzadeh et~al.(2019)Hasanzadeh, Hajiramezanali, Narayanan,
  Duffield, Zhou, and Qian]{hasanzadeh2019semi}
A.~Hasanzadeh, E.~Hajiramezanali, K.~Narayanan, N.~Duffield, M.~Zhou, and
  X.~Qian, ``Semi-implicit graph variational auto-encoders,'' in \emph{Advances
  in Neural Information Processing Systems}, vol.~32, 2019, pp.
  10\,712--10\,723.

\bibitem[Hu et~al.(2020{\natexlab{b}})Hu, Dong, Wang, Chang, and
  Sun]{hu2020gpt}
Z.~Hu, Y.~Dong, K.~Wang, K.-W. Chang, and Y.~Sun, ``{GPT-GNN}: Generative
  pre-training of graph neural networks,'' in \emph{Proceedings of the 26th ACM
  SIGKDD International Conference on Knowledge Discovery \& Data Mining}, 2020,
  pp. 1857--1867.

\bibitem[Peng et~al.(2020{\natexlab{b}})Peng, Dong, Luo, Wu, and
  Zheng]{peng2020self}
Z.~Peng, Y.~Dong, M.~Luo, X.-M. Wu, and Q.~Zheng, ``Self-supervised graph
  representation learning via global context prediction,'' \emph{arXiv preprint
  arXiv:2003.01604}, 2020.

\bibitem[Rong et~al.(2020)Rong, Bian, Xu, Xie, Wei, Huang, and
  Huang]{rong2020grover}
Y.~Rong, Y.~Bian, T.~Xu, W.~Xie, Y.~Wei, W.~Huang, and J.~Huang,
  ``Self-supervised graph transformer on large-scale molecular data,'' in
  \emph{Advances in Neural Information Processing Systems}, 2020.

\bibitem[Hwang et~al.(2020)Hwang, Park, Kwon, Kim, Ha, and Kim]{hwang2020self}
D.~Hwang, J.~Park, S.~Kwon, K.-M. Kim, J.-W. Ha, and H.~J. Kim,
  ``Self-supervised auxiliary learning with meta-paths for heterogeneous
  graphs,'' \emph{arXiv preprint arXiv:2007.08294}, 2020.

\bibitem[Sun et~al.(2020)Sun, Lin, and Zhu]{sun2020multi}
K.~Sun, Z.~Lin, and Z.~Zhu, ``Multi-stage self-supervised learning for graph
  convolutional networks on graphs with few labeled nodes,'' in
  \emph{Proceedings of the AAAI Conference on Artificial Intelligence},
  vol.~34, no.~04, 2020, pp. 5892--5899.

\bibitem[Hu et~al.(2021)Hu, Kou, Zhang, Li, Yang, and Liu]{hu2021rectifying}
Z.~Hu, G.~Kou, H.~Zhang, N.~Li, K.~Yang, and L.~Liu, ``Rectifying pseudo
  labels: Iterative feature clustering for graph representation learning,'' in
  \emph{Proceedings of the 30th ACM International Conference on Information \&
  Knowledge Management}, 2021, p. 720–729.

\bibitem[Zhang et~al.(2021)Zhang, Wu, Yan, Wipf, and
  Philip]{zhang2021canonical}
H.~Zhang, Q.~Wu, J.~Yan, D.~Wipf, and S.~Y. Philip, ``From canonical
  correlation analysis to self-supervised graph neural networks,'' in
  \emph{Advances in Neural Information Processing Systems}, 2021.

\bibitem[Xie et~al.(2022{\natexlab{a}})Xie, Xu, and Ji]{xie2022self}
Y.~Xie, Z.~Xu, and S.~Ji, ``Self-supervised representation learning via latent
  graph prediction,'' \emph{arXiv preprint arXiv:2202.08333}, 2022.

\bibitem[Liu et~al.(2020{\natexlab{d}})Liu, Zhang, Hou, Wang, Mian, Zhang, and
  Tang]{liu2020self}
X.~Liu, F.~Zhang, Z.~Hou, Z.~Wang, L.~Mian, J.~Zhang, and J.~Tang,
  ``Self-supervised learning: Generative or contrastive,'' \emph{arXiv preprint
  arXiv:2006.08218}, 2020.

\bibitem[Liu et~al.(2021{\natexlab{a}})Liu, Pan, Jin, Zhou, Xia, and
  Yu]{liu2021graph}
Y.~Liu, S.~Pan, M.~Jin, C.~Zhou, F.~Xia, and P.~S. Yu, ``Graph self-supervised
  learning: A survey,'' \emph{arXiv preprint arXiv:2103.00111}, 2021.

\bibitem[Wang et~al.(2021{\natexlab{a}})Wang, Zhang, Guo, Yin, Li, and
  Chen]{wang2021decoupling}
Y.~Wang, J.~Zhang, S.~Guo, H.~Yin, C.~Li, and H.~Chen, ``Decoupling
  representation learning and classification for gnn-based anomaly detection,''
  in \emph{Proceedings of the 44th International ACM SIGIR Conference on
  Research and Development in Information Retrieval}, 2021, p. 1239–1248.

\bibitem[Jin et~al.(2020)Jin, Derr, Liu, Wang, Wang, Liu, and
  Tang]{jin2020self}
W.~Jin, T.~Derr, H.~Liu, Y.~Wang, S.~Wang, Z.~Liu, and J.~Tang,
  ``Self-supervised learning on graphs: Deep insights and new direction,''
  \emph{arXiv preprint arXiv:2006.10141}, 2020.

\bibitem[Ganin and Lempitsky(2015)]{ganin2015unsupervised}
Y.~Ganin and V.~Lempitsky, ``Unsupervised domain adaptation by
  backpropagation,'' in \emph{International Conference on Machine
  Learning}.\hskip 1em plus 0.5em minus 0.4em\relax PMLR, 2015, pp. 1180--1189.

\bibitem[Sun et~al.(2019{\natexlab{b}})Sun, Tzeng, Darrell, and
  Efros]{sun2019unsupervised}
Y.~Sun, E.~Tzeng, T.~Darrell, and A.~A. Efros, ``Unsupervised domain adaptation
  through self-supervision,'' \emph{arXiv preprint arXiv:1909.11825}, 2019.

\bibitem[Hjelm et~al.(2019)Hjelm, Fedorov, Lavoie-Marchildon, Grewal, Bachman,
  Trischler, and Bengio]{hjelm2018learning}
R.~D. Hjelm, A.~Fedorov, S.~Lavoie-Marchildon, K.~Grewal, P.~Bachman,
  A.~Trischler, and Y.~Bengio, ``Learning deep representations by mutual
  information estimation and maximization,'' in \emph{International Conference
  on Learning Representations}, 2019.

\bibitem[Donsker and Varadhan(1983)]{donsker1983Asymptotic}
M.~D. Donsker and S.~R.~S. Varadhan, ``Asymptotic evaluation of certain markov
  process expectations for large time. iv,'' \emph{Communications on Pure and
  Applied Mathematics}, vol.~36, pp. 183--212, 1983.

\bibitem[Belghazi et~al.(2018)Belghazi, Baratin, Rajeshwar, Ozair, Bengio,
  Courville, and Hjelm]{belghazi2018mutual}
M.~I. Belghazi, A.~Baratin, S.~Rajeshwar, S.~Ozair, Y.~Bengio, A.~Courville,
  and D.~Hjelm, ``Mutual information neural estimation,'' in \emph{Proceedings
  of the 35th International Conference on Machine Learning}, vol.~80.\hskip 1em
  plus 0.5em minus 0.4em\relax PMLR, 2018, pp. 531--540.

\bibitem[Nowozin et~al.(2016)Nowozin, Cseke, and Tomioka]{nowozin2016f}
S.~Nowozin, B.~Cseke, and R.~Tomioka, ``f-gan: Training generative neural
  samplers using variational divergence minimization,'' in \emph{Advances in
  Neural Information Processing Systems}, 2016, pp. 271--279.

\bibitem[Gutmann and Hyv{\"a}rinen(2010)]{gutmann2010noise}
M.~Gutmann and A.~Hyv{\"a}rinen, ``Noise-contrastive estimation: A new
  estimation principle for unnormalized statistical models,'' in
  \emph{Proceedings of the Thirteenth International Conference on Artificial
  Intelligence and Statistics}, 2010, pp. 297--304.

\bibitem[Sohn(2016)]{sohn2016improved}
K.~Sohn, ``Improved deep metric learning with multi-class n-pair loss
  objective,'' in \emph{Advances in Neural Information Processing Systems},
  vol.~29, 2016, pp. 1857--1865.

\bibitem[Schroff et~al.(2015)Schroff, Kalenichenko, and
  Philbin]{schroff2015facenet}
F.~Schroff, D.~Kalenichenko, and J.~Philbin, ``Facenet: A unified embedding for
  face recognition and clustering,'' in \emph{Proceedings of the IEEE
  Conference on Computer Vision and Pattern Recognition}, 2015, pp. 815--823.

\bibitem[Hoffer and Ailon(2015)]{hoffer2015deep}
E.~Hoffer and N.~Ailon, ``Deep metric learning using triplet network,'' in
  \emph{International workshop on similarity-based pattern recognition}.\hskip
  1em plus 0.5em minus 0.4em\relax Springer, 2015, pp. 84--92.

\bibitem[Khosla et~al.(2020)Khosla, Teterwak, Wang, Sarna, Tian, Isola,
  Maschinot, Liu, and Krishnan]{khosla2020supervised}
P.~Khosla, P.~Teterwak, C.~Wang, A.~Sarna, Y.~Tian, P.~Isola, A.~Maschinot,
  C.~Liu, and D.~Krishnan, ``Supervised contrastive learning,'' \emph{Advances
  in Neural Information Processing Systems}, vol.~33, 2020.

\bibitem[Rendle et~al.(2009)Rendle, Freudenthaler, Gantner, and
  Schmidt-Thieme]{rendle2009bpr}
S.~Rendle, C.~Freudenthaler, Z.~Gantner, and L.~Schmidt-Thieme, ``Bpr: Bayesian
  personalized ranking from implicit feedback,'' in \emph{Proceedings of the
  Twenty-Fifth Conference on Uncertainty in Artificial Intelligence}, 2009, p.
  452–461.

\bibitem[Jiang et~al.(2021)Jiang, Jia, Fang, Shi, Lin, and Wang]{jiang2021pre}
X.~Jiang, T.~Jia, Y.~Fang, C.~Shi, Z.~Lin, and H.~Wang, ``Pre-training on
  large-scale heterogeneous graph,'' in \emph{Proceedings of the 27th ACM
  SIGKDD Conference on Knowledge Discovery and Data Mining}, 2021, pp.
  756–--766.

\bibitem[Wang et~al.(2021{\natexlab{b}})Wang, Liu, Han, and Shi]{xiao2021self}
X.~Wang, N.~Liu, H.~Han, and C.~Shi, ``Self-supervised heterogeneous graph
  neural network with co-contrastive learning,'' in \emph{Proceedings of the
  27th ACM SIGKDD Conference on Knowledge Discovery and Data Mining}, 2021, pp.
  1726--1736.

\bibitem[Tian et~al.(2020)Tian, Sun, Poole, Krishnan, Schmid, and
  Isola]{tian2020makes}
Y.~Tian, C.~Sun, B.~Poole, D.~Krishnan, C.~Schmid, and P.~Isola, ``What makes
  for good views for contrastive learning?'' \emph{arXiv preprint
  arXiv:2005.10243}, 2020.

\bibitem[Suresh et~al.(2021)Suresh, Li, Hao, and Neville]{suresh2021adversial}
S.~Suresh, P.~Li, C.~Hao, and J.~Neville, ``Adversarial graph augmentation to
  improve graph contrastive learning,'' \emph{Advances in Neural Information
  Processing Systems}, vol.~34, 2021.

\bibitem[Xu et~al.(2021)Xu, Cheng, Luo, Chen, and Zhang]{xu2021infogcl}
D.~Xu, W.~Cheng, D.~Luo, H.~Chen, and X.~Zhang, ``Infogcl: Information-aware
  graph contrastive learning,'' \emph{Advances in Neural Information Processing
  Systems}, vol.~34, 2021.

\bibitem[Wei et~al.(2021)Wei, Shen, Chen, and Ma]{wei2020theoretical}
C.~Wei, K.~Shen, Y.~Chen, and T.~Ma, ``Theoretical analysis of self-training
  with deep networks on unlabeled data,'' in \emph{International Conference on
  Learning Representations}, 2021.

\bibitem[Kramer(1991)]{kramer1991nonlinear}
M.~A. Kramer, ``Nonlinear principal component analysis using autoassociative
  neural networks,'' \emph{AIChE journal}, vol.~37, no.~2, pp. 233--243, 1991.

\bibitem[Hamilton(2020)]{book:hamilton}
W.~L. Hamilton, ``Graph representation learning,'' \emph{Synthesis Lectures on
  Artificial Intelligence and Machine Learning}, vol.~14, no.~3, pp. 1--159,
  2020.

\bibitem[Hamilton et~al.(2017)Hamilton, Ying, and
  Leskovec]{hamilton2017inductive}
W.~Hamilton, Z.~Ying, and J.~Leskovec, ``Inductive representation learning on
  large graphs,'' in \emph{Advances in Neural Information Processing Systems},
  2017, pp. 1024--1034.

\bibitem[Kim and Oh(2021)]{kim2021how}
D.~Kim and A.~Oh, ``How to find your friendly neighborhood: Graph attention
  design with self-supervision,'' in \emph{International Conference on Learning
  Representations}, 2021.

\bibitem[Jin et~al.(2021)Jin, Derr, Wang, Ma, Liu, and Tang]{jin2020node}
W.~Jin, T.~Derr, Y.~Wang, Y.~Ma, Z.~Liu, and J.~Tang, ``Node similarity
  preserving graph convolutional networks,'' in \emph{Proceedings of the 14th
  ACM International Conference on Web Search and Data Mining}.\hskip 1em plus
  0.5em minus 0.4em\relax ACM, 2021.

\bibitem[Vincent et~al.(2010)Vincent, Larochelle, Lajoie, Bengio, Manzagol, and
  Bottou]{vincent2010stacked}
P.~Vincent, H.~Larochelle, I.~Lajoie, Y.~Bengio, P.-A. Manzagol, and L.~Bottou,
  ``Stacked denoising autoencoders: Learning useful representations in a deep
  network with a local denoising criterion.'' \emph{Journal of Machine Learning
  Research}, vol.~11, no.~12, 2010.

\bibitem[Taubin(1995)]{taubin1995signal}
G.~Taubin, ``A signal processing approach to fair surface design,'' in
  \emph{Proceedings of the 22nd annual conference on Computer graphics and
  interactive techniques}, 1995, pp. 351--358.

\bibitem[You et~al.(2020{\natexlab{b}})You, Chen, Wang, and Shen]{you2020does}
Y.~You, T.~Chen, Z.~Wang, and Y.~Shen, ``When does self-supervision help graph
  convolutional networks?'' in \emph{International Conference on Machine
  Learning}.\hskip 1em plus 0.5em minus 0.4em\relax PMLR, 2020, pp.
  10\,871--10\,880.

\bibitem[Goodfellow et~al.(2014)Goodfellow, Pouget-Abadie, Mirza, Xu,
  Warde-Farley, Ozair, Courville, and Bengio]{goodfellow2014generative}
I.~Goodfellow, J.~Pouget-Abadie, M.~Mirza, B.~Xu, D.~Warde-Farley, S.~Ozair,
  A.~Courville, and Y.~Bengio, ``Generative adversarial nets,'' in
  \emph{Advances in Neural Information Processing Systems}, vol.~27, 2014, pp.
  2672--2680.

\bibitem[Jin et~al.(2018)Jin, Barzilay, and Jaakkola]{jin2018junction}
W.~Jin, R.~Barzilay, and T.~Jaakkola, ``Junction tree variational autoencoder
  for molecular graph generation,'' in \emph{International Conference on
  Machine Learning}.\hskip 1em plus 0.5em minus 0.4em\relax PMLR, 2018, pp.
  2323--2332.

\bibitem[Li et~al.(2020)Li, Li, Zhang, Zhao, Rong, and Cheng]{li2020dirichlet}
J.~Li, T.~Y.~J. Li, H.~Zhang, K.~Zhao, Y.~Rong, and H.~Cheng, ``Dirichlet graph
  variational autoencoder,'' in \emph{Advances in Neural Information Processing
  Systems}, 2020.

\bibitem[Radford et~al.(2019)Radford, Wu, Child, Luan, Amodei, and
  Sutskever]{radford2019language}
A.~Radford, J.~Wu, R.~Child, D.~Luan, D.~Amodei, and I.~Sutskever, ``Language
  models are unsupervised multitask learners,'' \emph{OpenAI blog}, vol.~1,
  no.~8, p.~9, 2019.

\bibitem[Hotelling(1992)]{hotelling1992relations}
H.~Hotelling, ``Relations between two sets of variates,'' in
  \emph{Breakthroughs in statistics}.\hskip 1em plus 0.5em minus 0.4em\relax
  Springer, 1992, pp. 162--190.

\bibitem[Hardoon et~al.(2004)Hardoon, Szedmak, and
  Shawe-Taylor]{hardoon2004canonical}
D.~R. Hardoon, S.~Szedmak, and J.~Shawe-Taylor, ``Canonical correlation
  analysis: An overview with application to learning methods,'' \emph{Neural
  computation}, vol.~16, no.~12, pp. 2639--2664, 2004.

\bibitem[Zbontar et~al.(2021)Zbontar, Jing, Misra, LeCun, and
  Deny]{zbontar2021barlow}
J.~Zbontar, L.~Jing, I.~Misra, Y.~LeCun, and S.~Deny, ``Barlow twins:
  Self-supervised learning via redundancy reduction,'' in \emph{International
  Conference on Machine Learning}.\hskip 1em plus 0.5em minus 0.4em\relax PMLR,
  2021, pp. 12\,310--12\,320.

\bibitem[Tishby et~al.(1999)Tishby, Pereira, and Bialek]{tishby99information}
N.~Tishby, F.~C. Pereira, and W.~Bialek, ``The information bottleneck method,''
  in \emph{Proceedings of the 37-th Annual Allerton Conference on
  Communication, Control and Computing}, 1999, pp. 368--377.

\bibitem[Li et~al.(2018)Li, Han, and Wu]{li2018deeper}
Q.~Li, Z.~Han, and X.-M. Wu, ``Deeper insights into graph convolutional
  networks for semi-supervised learning,'' in \emph{Proceedings of the AAAI
  Conference on Artificial Intelligence}, vol.~32, no.~1, 2018.

\bibitem[Caron et~al.(2018)Caron, Bojanowski, Joulin, and Douze]{caron2018deep}
M.~Caron, P.~Bojanowski, A.~Joulin, and M.~Douze, ``Deep clustering for
  unsupervised learning of visual features,'' in \emph{Proceedings of the
  European Conference on Computer Vision}, 2018, pp. 132--149.

\bibitem[Morris et~al.(2020)Morris, Kriege, Bause, Kersting, Mutzel, and
  Neumann]{Morris2020tudataset}
C.~Morris, N.~M. Kriege, F.~Bause, K.~Kersting, P.~Mutzel, and M.~Neumann,
  ``Tudataset: A collection of benchmark datasets for learning with graphs,''
  in \emph{ICML 2020 Workshop on Graph Representation Learning and Beyond},
  2020.

\bibitem[{Wale} and {Karypis}(2006)]{NCI1}
N.~{Wale} and G.~{Karypis}, ``Comparison of descriptor spaces for chemical
  compound retrieval and classification,'' in \emph{Sixth International
  Conference on Data Mining}, 2006, pp. 678--689.

\bibitem[Debnath et~al.(1991)Debnath, Lopez~de Compadre, Debnath, Shusterman,
  and Hansch]{mutag}
A.~K. Debnath, R.~L. Lopez~de Compadre, G.~Debnath, A.~J. Shusterman, and
  C.~Hansch, ``Structure-activity relationship of mutagenic aromatic and
  heteroaromatic nitro compounds correlation with molecular orbital energies
  and hydrophobicity,'' \emph{Journal of Medicinal Chemistry}, vol.~34, no.~2,
  pp. 786--797, 1991.

\bibitem[Sterling and Irwin(2015)]{sterling2015zinc}
T.~Sterling and J.~J. Irwin, ``Zinc 15--ligand discovery for everyone,''
  \emph{Journal of Chemical Information and Modeling}, vol.~55, no.~11, pp.
  2324--2337, 2015.

\bibitem[Borgwardt et~al.(2005)Borgwardt, Ong, Schönauer, Vishwanathan, Smola,
  and Kriegel]{proteins}
K.~M. Borgwardt, C.~S. Ong, S.~Schönauer, S.~V.~N. Vishwanathan, A.~J. Smola,
  and H.-P. Kriegel, ``Protein function prediction via graph kernels,''
  \emph{Bioinformatics}, vol.~21, pp. i47--i56, 2005.

\bibitem[Dobson and Doig(2003)]{DD}
P.~D. Dobson and A.~J. Doig, ``Distinguishing enzyme structures from
  non-enzymes without alignments,'' \emph{Journal of Molecular Biology}, vol.
  330, no.~4, pp. 771--783, 2003.

\bibitem[Yanardag and Vishwanathan(2015)]{deep2015yanardag}
P.~Yanardag and S.~Vishwanathan, ``Deep graph kernels,'' in \emph{Proceedings
  of the 21th ACM SIGKDD International Conference on Knowledge Discovery and
  Data Mining}, 2015, p. 1365–1374.

\bibitem[Yang et~al.(2016)Yang, Cohen, and Salakhudinov]{yang2016revisiting}
Z.~Yang, W.~Cohen, and R.~Salakhudinov, ``Revisiting semi-supervised learning
  with graph embeddings,'' in \emph{International conference on machine
  learning}.\hskip 1em plus 0.5em minus 0.4em\relax PMLR, 2016, pp. 40--48.

\bibitem[McCallum et~al.(2000)McCallum, Nigam, Rennie, and
  Seymore]{mccallum2000cora}
A.~K. McCallum, K.~Nigam, J.~Rennie, and K.~Seymore, ``Automating the
  construction of internet portals with machine learning,'' \emph{Information
  Retrieval}, vol.~3, no.~2, pp. 127--163, 2000.

\bibitem[Giles et~al.(1998)Giles, Bollacker, and Lawrence]{citeseer}
C.~L. Giles, K.~D. Bollacker, and S.~Lawrence, ``Citeseer: An automatic
  citation indexing system,'' in \emph{Proceedings of the Third ACM Conference
  on Digital Libraries}.\hskip 1em plus 0.5em minus 0.4em\relax Association for
  Computing Machinery, 1998, p. 89–98.

\bibitem[Sen et~al.(2008)Sen, Namata, Bilgic, Getoor, Galligher, and
  Eliassi-Rad]{sen2008pubmed}
P.~Sen, G.~Namata, M.~Bilgic, L.~Getoor, B.~Galligher, and T.~Eliassi-Rad,
  ``Collective classification in network data,'' \emph{AI magazine}, vol.~29,
  no.~3, pp. 93--93, 2008.

\bibitem[Shchur et~al.(2018)Shchur, Mumme, Bojchevski, and
  G{\"u}nnemann]{shchur2018pitfalls}
O.~Shchur, M.~Mumme, A.~Bojchevski, and S.~G{\"u}nnemann, ``Pitfalls of graph
  neural network evaluation,'' \emph{arXiv preprint arXiv:1811.05868}, 2018.

\bibitem[Sinha et~al.(2015)Sinha, Shen, Song, Ma, Eide, Hsu, and
  Wang]{sinha2015overview}
A.~Sinha, Z.~Shen, Y.~Song, H.~Ma, D.~Eide, B.-J. Hsu, and K.~Wang, ``An
  overview of microsoft academic service ({MAS}) and applications,'' in
  \emph{Proceedings of the 24th international conference on world wide web},
  2015, pp. 243--246.

\bibitem[McAuley et~al.(2015)McAuley, Targett, Shi, and Van
  Den~Hengel]{mcauley2015image}
J.~McAuley, C.~Targett, Q.~Shi, and A.~Van Den~Hengel, ``Image-based
  recommendations on styles and substitutes,'' in \emph{Proceedings of the 38th
  international ACM SIGIR conference on research and development in information
  retrieval}, 2015, pp. 43--52.

\bibitem[Zeng et~al.(2020)Zeng, Zhou, Srivastava, Kannan, and
  Prasanna]{graphsaint-iclr20}
H.~Zeng, H.~Zhou, A.~Srivastava, R.~Kannan, and V.~Prasanna, ``{GraphSAINT}:
  Graph sampling based inductive learning method,'' in \emph{International
  Conference on Learning Representations}, 2020.

\bibitem[Zitnik and Leskovec(2017)]{zitnik2017predicting}
M.~Zitnik and J.~Leskovec, ``Predicting multicellular function through
  multi-layer tissue networks,'' \emph{Bioinformatics}, vol.~33, no.~14, pp.
  i190--i198, 2017.

\bibitem[Zitnik et~al.(2019)Zitnik, Feldman, Leskovec,
  et~al.]{zitnik2019evolution}
M.~Zitnik, M.~W. Feldman, J.~Leskovec \emph{et~al.}, ``Evolution of resilience
  in protein interactomes across the tree of life,'' \emph{Proceedings of the
  National Academy of Sciences}, vol. 116, no.~10, pp. 4426--4433, 2019.

\bibitem[Liu et~al.(2021{\natexlab{b}})Liu, Luo, Wang, Xie, Yuan, Gui, Yu, Xu,
  Zhang, Liu, Yan, Liu, Fu, Oztekin, Zhang, and Ji]{liu2021dig}
M.~Liu, Y.~Luo, L.~Wang, Y.~Xie, H.~Yuan, S.~Gui, H.~Yu, Z.~Xu, J.~Zhang,
  Y.~Liu, K.~Yan, H.~Liu, C.~Fu, B.~M. Oztekin, X.~Zhang, and S.~Ji, ``{DIG}: A
  turnkey library for diving into graph deep learning research,'' \emph{Journal
  of Machine Learning Research}, vol.~22, no. 240, pp. 1--9, 2021.

\bibitem[Paszke et~al.(2019)Paszke, Gross, Massa, Lerer, Bradbury, Chanan,
  Killeen, Lin, Gimelshein, Antiga, Desmaison, Kopf, Yang, DeVito, Raison,
  Tejani, Chilamkurthy, Steiner, Fang, Bai, and Chintala]{Paszke2019Pytorch}
A.~Paszke, S.~Gross, F.~Massa, A.~Lerer, J.~Bradbury, G.~Chanan, T.~Killeen,
  Z.~Lin, N.~Gimelshein, L.~Antiga, A.~Desmaison, A.~Kopf, E.~Yang, Z.~DeVito,
  M.~Raison, A.~Tejani, S.~Chilamkurthy, B.~Steiner, L.~Fang, J.~Bai, and
  S.~Chintala, ``Pytorch: An imperative style, high-performance deep learning
  library,'' in \emph{Advances in Neural Information Processing Systems 32},
  2019, pp. 8024--8035.

\bibitem[Fey and Lenssen(2019)]{Fey2019Geometric}
M.~Fey and J.~E. Lenssen, ``Fast graph representation learning with {PyTorch
  Geometric},'' in \emph{ICLR Workshop on Representation Learning on Graphs and
  Manifolds}, 2019.

\bibitem[You et~al.(2018)You, Ying, Ren, Hamilton, and
  Leskovec]{you2018graphrnn}
J.~You, R.~Ying, X.~Ren, W.~Hamilton, and J.~Leskovec, ``{G}raph{RNN}:
  Generating realistic graphs with deep auto-regressive models,'' in
  \emph{Proceedings of the 35th International Conference on Machine Learning},
  vol.~80.\hskip 1em plus 0.5em minus 0.4em\relax PMLR, 2018, pp. 5708--5717.

\bibitem[Shi* et~al.(2020)Shi*, Xu*, Zhu, Zhang, Zhang, and
  Tang]{Shi2020GraphAF:}
C.~Shi*, M.~Xu*, Z.~Zhu, W.~Zhang, M.~Zhang, and J.~Tang, ``Graphaf: a
  flow-based autoregressive model for molecular graph generation,'' in
  \emph{International Conference on Learning Representations}, 2020.

\bibitem[Zang and Wang(2020)]{zhang2020moflow}
C.~Zang and F.~Wang, ``Moflow: An invertible flow model for generating
  molecular graphs,'' in \emph{Proceedings of the 26th ACM SIGKDD International
  Conference on Knowledge Discovery and Data Mining}, 2020, p. 617–626.

\bibitem[Luo et~al.(2021)Luo, Yan, and Ji]{luo2021graphdf}
Y.~Luo, K.~Yan, and S.~Ji, ``Graphdf: A discrete flow model for molecular graph
  generation,'' in \emph{Proceedings of the 38th International Conference on
  Machine Learning}, vol. 139.\hskip 1em plus 0.5em minus 0.4em\relax PMLR,
  2021, pp. 7192--7203.

\bibitem[Li et~al.(2021)Li, Wang, Li, Qiao, Liu, Ma, Gao, Song, and
  Xie]{li2021pairwise}
P.~Li, J.~Wang, Z.~Li, Y.~Qiao, X.~Liu, F.~Ma, P.~Gao, S.~Song, and G.~Xie,
  ``Pairwise half-graph discrimination: A simple graph-level self-supervised
  strategy for pre-training graph neural networks,'' in \emph{Proceedings of
  the Thirtieth International Joint Conference on Artificial Intelligence},
  2021, pp. 2694--2700.

\bibitem[Yuan et~al.(2020{\natexlab{b}})Yuan, Yu, Gui, and
  Ji]{yuan2020explainability}
H.~Yuan, H.~Yu, S.~Gui, and S.~Ji, ``Explainability in graph neural networks: A
  taxonomic survey,'' \emph{arXiv preprint arXiv:2012.15445}, 2020.

\bibitem[Xie et~al.(2022{\natexlab{b}})Xie, Katariya, Tang, Huang, Rao,
  Subbian, and Ji]{xie2022task}
Y.~Xie, S.~Katariya, X.~Tang, E.~Huang, N.~Rao, K.~Subbian, and S.~Ji,
  ``Task-agnostic graph explanations,'' \emph{arXiv preprint arXiv:2202.08335},
  2022.

\bibitem[Schlichtkrull et~al.(2018)Schlichtkrull, Kipf, Bloem, Berg, Titov, and
  Welling]{schlichtkrull2018modeling}
M.~Schlichtkrull, T.~N. Kipf, P.~Bloem, R.~v.~d. Berg, I.~Titov, and
  M.~Welling, ``Modeling relational data with graph convolutional networks,''
  in \emph{European semantic web conference}.\hskip 1em plus 0.5em minus
  0.4em\relax Springer, 2018, pp. 593--607.

\bibitem[Tian et~al.(2021)Tian, Wu, Shi, Zhu, and Xiong]{tian2021self}
S.~Tian, R.~Wu, L.~Shi, L.~Zhu, and T.~Xiong, ``Self-supervised representation
  learning on dynamic graphs,'' in \emph{Proceedings of the 30th ACM
  International Conference on Information \& Knowledge Management}, 2021, pp.
  1814--1823.

\bibitem[Veličković et~al.(2018)Veličković, Cucurull, Casanova, Romero,
  Liò, and Bengio]{vel2018graph}
P.~Veličković, G.~Cucurull, A.~Casanova, A.~Romero, P.~Liò, and Y.~Bengio,
  ``Graph attention networks,'' in \emph{International Conference on Learning
  Representations}, 2018.

\bibitem[Xu et~al.(2018)Xu, Li, Tian, Sonobe, Kawarabayashi, and
  Jegelka]{xu2018representation}
K.~Xu, C.~Li, Y.~Tian, T.~Sonobe, K.-i. Kawarabayashi, and S.~Jegelka,
  ``Representation learning on graphs with jumping knowledge networks,'' in
  \emph{International Conference on Machine Learning}.\hskip 1em plus 0.5em
  minus 0.4em\relax PMLR, 2018, pp. 5453--5462.

\bibitem[Erhan et~al.(2010)Erhan, Bengio, Courville, Manzagol, Vincent, and
  Bengio]{erhan2010why}
D.~Erhan, Y.~Bengio, A.~Courville, P.-A. Manzagol, P.~Vincent, and S.~Bengio,
  ``Why does unsupervised pre-training help deep learning?'' \emph{Journal of
  Machine Learning Research}, vol.~11, no.~19, pp. 625--660, 2010.

\bibitem[Tschannen et~al.(2020)Tschannen, Djolonga, Rubenstein, Gelly, and
  Lucic]{Tschannen2020On}
M.~Tschannen, J.~Djolonga, P.~K. Rubenstein, S.~Gelly, and M.~Lucic, ``On
  mutual information maximization for representation learning,'' in
  \emph{International Conference on Learning Representations}, 2020.

\bibitem[Mikolov et~al.(2013)Mikolov, Chen, Corrado, and
  Dean]{tomas2013efficient}
T.~Mikolov, K.~Chen, G.~Corrado, and J.~Dean, ``Efficient estimation of word
  representations in vector space,'' in \emph{1st International Conference on
  Learning Representations, Workshop Track Proceedings}, 2013.

\end{thebibliography}

\newpage
\appendices 
\section{Heterogeneous and Dynamic Graphs}
\newrevision{Compared to typical homogeneous graphs, the heterogeneous graphs further include attributes indicating types of nodes and edges that contain richer topological and feature information. 
Generally, typical contrastive and predictive frameworks can still be adapted to learn representations from heterogeneous and dynamic graphs, as long as a proper graph encoder, such as message passing neural networks with edge attributes and R-GCNs~\cite{schlichtkrull2018modeling}. To better utilize the information in node and edge types, recent work proposes contrastive methods with view generations and objectives specifically designed for heterogeneous graphs. In particular, for the view generation, HeCo~\cite{xiao2021self} proposes to generate the network schema and meta-paths as two views in the contrastive framework for heterogeneous graphs, as introduced in Section 3.3.3. Moreover, for the computation of contrastive objectives, HGNN~\cite{jiang2021pre} proposes to adopt asymmetric projection heads for the representation of two nodes with different types. For predictive methods, the meta-paths in heterogeneous graphs are adopted as additional self-supervision to help improve the performance of downstream tasks. The dynamic graph can be considered as a special case of heterogeneous where the additional attributes contain temporal information indicating the time when the nodes and edges are constructed in a continuous form. \citet{tian2021self} propose the time-aware GNN encoder for dynamic graphs and the DDGCL framework where two temporal subgraphs obtained at different time points are adopted as two views. While still following the general contrastive framework, the specific design of view generation, encoders, and objectives for homogeneous and dynamic graphs can bring additional performance gain compared to general contrastive methods on homogeneous graphs.}

\section{Comparisons between Contrastive and Predictive Models}

\newrevision{As described in Section 1, the major methodological difference between contrastive methods and predictive methods is whether paired samples are required for training, as contrastive methods contrast negative pairs from positive ones. They both aim to learn encoders that compute informative representations. For contrastive learning, the goal is achieved by maximizing mutual information between representations of different parts of the data. In other words, the mutual information $I(v_i, f(v_i))$ between any given view $v_i$ and its representation $f(v_i)$ is maximized only if $I(f(v_i), f(v_j))$ is maximized, ideally, to $I(v_i, v_j)$ for any views of a given graph. For predictive methods, the goal is achieved by learning representations that preserve (by being able to predict) certain properties or characteristics of the original graph.}

\newrevision{Empirically, contrastive methods are usually more computationally expensive compared to predictive methods but generally outperform most predictive methods in terms of downstream classification performance. On the other hand, recent predictive methods based on invariance-regularization can achieve performance on par with the SOTA contrastive methods.
The design of pretext learning tasks in predictive methods is more flexible so more domain knowledge can be included to benefit the learning of representations. However, most predictive methods are less theoretically guided compared to contrastive methods grounded on mutual information, as discussed in Section 7, which may lead to the reduced performance mentioned above.}

\section{Graph Encoders in Contrastive Learning}

Graph encoders are usually constructed based on graph neural networks (GNNs) following a neighborhood aggregation strategy, where the representation of a node is iteratively updated by combining the its own representation with the aggregated representation over its neighbors. Formally, the $k$-th layer of GNN is:
\begin{equation}
    \bm x_v^{(k)} = \mbox{COMBINE}^{(k)} (\bm x_v^{(k-1)}, \bm a_v^k),
\end{equation}
\begin{equation}
    \bm a_v^k = \mbox{AGGREGATE}^{(k)} \left ({(\bm x_v^{(k-1)}, \bm x_u^{(k-1)}): u \in \mathcal{N}(v)} \right ),
\end{equation}
where $\bm x_v^{(k)}$ denotes the feature vector of node $v$ at the $k$-th layer, and $\mathcal{N}(v)$ is a set of neighbor nodes of $v$. Graph encoders mainly differ from their aggregation strategies. $\mbox{COMBINE}^{(k)}(\cdot)$ and $\mbox{AGGREGATE}^{(k)}(\cdot)$ are component functions that determine types of GNNs, such as Graph Convolutional Networks (GCNs)~\cite{kipf2017semi}, Graph Attention Networks (GATs)~\cite{vel2018graph} and Graph Isomorphism Networks (GINs)~\cite{xu2018how}.

\subsection{Node-Level and Graph-Level Representations}
The most straight-forward way to obtain the node-level representation $\bm h_v$ for node $v$ is to directly use the node feature at the final layer $K$ of the encoder~\cite{velikovi2019deep, Hu2020Strategies, jiao2020sub}, \emph{i.e.}, $\bm h_v = \bm x_v^{(K)}$. One may also adopt skip connections or jumping knowledge~\cite{xu2018representation} to generate node-level representation. However, the node-level representation produced by concatenating node features from all layers have different dimension from node features. To avoid such inconsistency in vector dimension,
% \cite{sun2019infograph} and \cite{you2020graph} concatenate node features from all layers as:
% \begin{equation}
%     \bm h_v = CONCAT([\bm x_v^{(k)}]_{k=1}^K),
% \end{equation}
% where $\bm h_v$ and caused global representation $\bm h_{graph}$ have different dimensions from node features. To avoid such inconsistency in vector dimension,
\cite{sun2019infograph, you2020graph} and \cite{hassani2020contrastive} concatenate node features of all layers, followed by a linear transformation:
\begin{equation}
    \bm h_v = \mbox{CONCAT}([\bm x_v^{(k)}]_{k=1}^K) \bm W,
\end{equation}
where $\bm W \in \mathbb{R}^{(\sum_k d_k) \times d}$ is the weight matrix used to shrink the dimension size of $\bm h_v $.

The $\mbox{READOUT}$ function is considered as the key operation to compute the graph-level representation $\bm h_{graph}$ given the node-level representations $\bm H$ of the graph. For the sake of node permutation invariance, summation and averaging are most commonly used READOUT functions. \citet{sun2019infograph, you2020graph} and \citet{hassani2020contrastive} employ sum over all the nodes' representations as
\begin{equation}
    \bm h_{graph} = \mbox{READOUT}(\bm H) = \sigma (\sum_{v=1}^{|V|} \bm h_v),
\end{equation}
where $|V|$ denotes the total number of nodes in the given graph, and $\sigma$ is either sigmoid function, multi-layer perceptron or identity function. \citet{velikovi2019deep} and \citet{jiao2020sub} employ mean pooling READOUT that averages all the node-level representations as
\begin{equation}
    \bm h_{graph} = \mbox{READOUT}(\bm H) = \sigma (\frac{1}{|V|} \sum_{v=1}^{|V|} \bm h_v).
\end{equation}

\subsection{Effects of Graph Encoders}

Typically, GNN-based encoders are not constrained on choices of GNN types and most frameworks~\cite{hassani2020contrastive, qiu2020gcc} allow various choices. However, some studies have more thorough considerations of GNN types. InfoGraph~\cite{sun2019infograph} adopts GIN to achieve less inductive bias for graph-level applications. GraphCL~\cite{you2020graph} finds GIN outperforms GCN and GAT in semi-supervised learning on node classification tasks. \citet{Hu2020Strategies} observe that the most expressive GIN achieves the best performance with pre-training, although GIN has slightly inferior performance than the less expressive GNNs without pre-training. That is, GIN achieves the highest performance gain of pre-training. This observation agrees with \cite{erhan2010why} that fully-utilization of pre-training requires an expressive model as limited expressive models can harm performance and the observation~\cite{Tschannen2020On} that the quality of learned representations is impacted more by the choice of encoder than the objective. On the contrary, for SUBG-CON~\cite{jiao2020sub}, GCN-based encoders outperform other GNN-based encoders as GCN is more suitable to handle subgraphs than more expressive GIN and GAT. In addition, \citet{velikovi2019deep} and \citet{zhu2020deep} employ different encoders on different learning tasks, \emph{i.e.}, GCN for transductive learning tasks, GraphSage-GCN or a mean-pooling layer with skip connection for inductive learning on larges Reddit, and mean-pooling layers with skip or dense skip connections for inductive learning on multiple graphs PPI. These observations imply that different contrastive learning frameworks and methods may prefer distinct GNN types for encoders. Even for the same framework, encoder choices may vary when applying to different datasets.

\section{Comparison of Contrastive Objectives}

Among all contrastive objectives for graphs, the JS estimator $\mathcal{\widehat I}^{(JS)}$ and InfoNCE $\mathcal{\widehat I}^{(\mathrm{NCE})}$ based on lower-bounds to mutual information are most commonly used. Regarding the two estimators generally, \citet{hjelm2018learning} empirically shows that InfoNCE generally outperforms the JS estimator in most cases. However, compared to the JS estimator, InfoNCE is more sensitive to the number of negative samples $N$ and requires a large number of negative samples in order to be competitive. Consequently, when the number of negative samples, \emph{i.e.}, the mini-batch size, is limited, the performance of InfoNCE could be limited and the JS estimator may become a better choice.

In addition, \citet{hassani2020contrastive} perform ablation studies comparing the three objectives $\mathcal{\widehat I}^{(\mathrm{DV})}, \mathcal{\widehat I}^{(\mathrm{JS})}$ and $\mathcal{\widehat I}^{(\mathrm{NCE})}$ with batch size ranged from $32$ to $256$ for graph classification and ranged from $2$ to $8$ for node classification. They show that the JS estimator generally leads to the best performance among all objectives on graph classification datasets, while InfoNCE (or NT-Xent) achieves overall best performance on node-classification tasks.

Moreover, \citet{jiao2020sub} compares the non-bound based triplet margin loss, the logistic loss~\cite{tomas2013efficient} as an equivalence of the JS estimator and the BPR loss as an equivalence of the InfoNCE with $N=1$ under their graph contrastive framework. Their results show that the JS estimator and the BPR loss (the InfoNCE loss with $N=1$) are similarly effective in their method, while the triplet margin loss achieves the best performance among the three objectives. The results indicate that the triplet margin loss can still be effective, given some certain views of the graphs, when the positive pairs and negative pairs should not be discriminated absolutely.

\section{Summary of SSL Methods for GNNs}
We summarize contrastive methods and predictive methods reviewed in this survey in Supplementary Table~1 and Supplementary Table~2, respectively.

\begin{table*}[h]\label{tab:s1}
\centering
\revision{\caption{Summary of contrastive methods for GNNs in chronological order. Columns categorize the methods by their learning objectives, level of representations for contrast (G: graph, N: node), the type of view generation, and the level of their majorly targeted downstream tasks. We also include notes to show their specific constraints or related literature in other domains such as vision for additional references, when applicable. For downstream tasks, the task of link prediction is also considered as node-level as it is based on representations of a pair of nodes. *The triplet loss is equivalent to the NT-Xent (InfoNCE) loss where the number of negative samples equals to one.}
\begin{tabular}{lccccc}
\toprule
Method    & Objective  & Rep. Levels & View Generation & Targeted Downstream Tasks   & Other Notes \\\hline
DGI~\cite{velikovi2019deep}         & JSE        & G-N         & Identical       & Node-level               & Deep Infomax~\cite{hjelm2018learning} \\\hline
InfoGraph~\cite{sun2019infograph}   & JSE        & G-N         & Identical       & Graph-level              & Deep Infomax~\cite{hjelm2018learning} \\\hline
\citet{Hu2020Strategies}            & JSE        & G-G         & Subgraphs       & Graph-level              & --                    \\\hline
GMI~\cite{peng2020graph}            & JSE        & N-N         & Identical       & Node-level               & --                    \\\hline
GCC~\cite{qiu2020gcc}               & InfoNCE    & G-G         & Subgraphs       & Graph-level              & --                    \\\hline
SUBG-CON~\cite{jiao2020sub}         & Triplet*   & G-G         & Subgraphs       & Graph-level              & --                    \\\hline
GRACE~\cite{zhu2020deep}            & InfoNCE    & N-N         & Structural \& Feature & Node-level         & Molecular Graph       \\\hline
MVGRL~\cite{hassani2020contrastive} & JSE        & G-N         & Structural \& Subgraphs & Node-level       & --                    \\\hline
GraphCL~\cite{you2020graph}         & InfoNCE    & G-G         & Random          & Graph-level              & SimCLR~\cite{chen2020simple}  \\\hline
GCA~\cite{zhu2021graph}             & InfoNCE    & N-N         & Structural \& Subgraphs & Node-level               & --                    \\\hline
PHD~\cite{li2021pairwise}           & JSE        & G-G         & Subgraphs       & Graph-level              & Molecular Graph    \\\hline
PT-HGNN~\cite{jiang2021pre}         & InfoNCE    & N-N         & Structural      & Node-level               & Heterogeneous Graph\\\hline
HeCo~\cite{xiao2021self}            & InfoNCE    & N-N         & \begin{tabular}[c]{@{}c@{}}Subgraphs \\ (Schema \& Meta-path)\end{tabular} & Node-level & Heterogeneous Graph\\\hline
InfoGCL~\cite{xu2021infogcl}        & InfoNCE    & G-G/G-N/N-N & Optimized (searched)  & Graph-level/Node-level   & Tian et al.~\cite{tian2020makes}        \\\hline
AD-GCL~\cite{suresh2021adversial}   & InfoNCE    & G-G         & Optimized (learned)   & Graph-level              & Tian et al.~\cite{tian2020makes}         \\
\bottomrule
\end{tabular}}
\end{table*}

\begin{table*}[h]\label{tab:s2}
\centering
\revision{\caption{Summary of predictive methods for GNNs. Columns categorize the methods by their sources of supervision, sub-categories, pretext tasks, and paradigms of utilizing self-supervision. In the training paradigm column, \textit{URL} denotes unsupervised representation learning, \textit{Pretrain} denotes unsupervised pretraining, and \textit{Auxiliary} denotes auxiliary learining.}
\begin{tabular}{llllc}
\toprule
Method       & Source of Supervision             & Sub-category                                                                               & Pretext Task                        & Training Paradigm \\\hline
GAE~\cite{kipf2016variational}          & \multirow{9}{*}{Reconstruction}   & Graph autoencoder                                                                          & Adjacency reconstruction            & URL               \\
VGAE~\cite{kipf2016variational}         &                                   & Variational autoencoder                                                                    & Adjacency reconstruction            & URL               \\
MGAE~\cite{wang2017mgae}         &                                   & Denoising autoencoder                                                                      & Node feature reconstruction         & URL               \\
ARGA/ARVGA~\cite{pan2018adversarially}   &                                   & Variational autoencoder                                                                    & Adjacency reconstruction            & URL               \\
GALA~\cite{park2019symmetric}         &                                   & Graph autoencoder                                                                          & Node feature reconstruction         & URL               \\
SIG-VGA~\cite{hasanzadeh2019semi}      &                                   & Variational autoencoder                                                                    & Adjacency reconstruction            & URL               \\
GPT-GNN~\cite{hu2020gpt}      &                                   & Auto-regressive reconstruction                                                             & Node and edge reconstruction        & URL               \\
SuperGAT~\cite{kim2021how}     &                                   & Graph autoencoder                                                                          & Adjacency reconstruction            & Auxiliary         \\
SimP-GCN~\cite{jin2020node}     &                                   & Graph autoencoder                                                                          & Node-pair similarity rec. & Auxiliary         \\\hline
BGRL~\cite{thakoor2021bootstrapped}             & \multirow{3}{*}{\begin{tabular}[c]{@{}l@{}}Invariance \\ regularization\end{tabular}}   & --        & Pseudo-contrastive            & URL               \\
CCA-SSG~\cite{zhang2021canonical}               &   & --        & Correlation reduction                 & URL               \\
LaGraph~\cite{xie2022self}      &   & --        & Latent graph prediction               & URL               \\\hline
$S^2$GRL~\cite{peng2020self}        & \multirow{4}{*}{Graph properties} & Statistical property                                                            & K-hop connectivity prediction       & URL               \\
GROVER~\cite{rong2020grover}       &                                   & \begin{tabular}[c]{@{}l@{}}Statistical and domain \\ knowledge-based\end{tabular} & Motif, contextual property pred.     & Pretrain          \\
\citet{hwang2020self} &                                   & Topological property                                                            & Meta-path prediction                & Auxiliary         \\\hline
M3S~\cite{sun2020multi}          & \multirow{2}{*}{Pseudo-labels}   & Self-training  & Pseudo-label prediction             & URL           \\   
IFC-GCN~\cite{hu2021rectifying}      &   & Self-training  & Pseudo-label prediction             & Pretrain          
\\\bottomrule
\end{tabular}}
\end{table*}

\section{Overview of the SSLGraph library within DIG}
An overview of the developed DIG-sslgraph library is shown in Supplementary Figure~1.

\begin{figure*}[ht]
    \centering
    \includegraphics[width=0.95\textwidth]{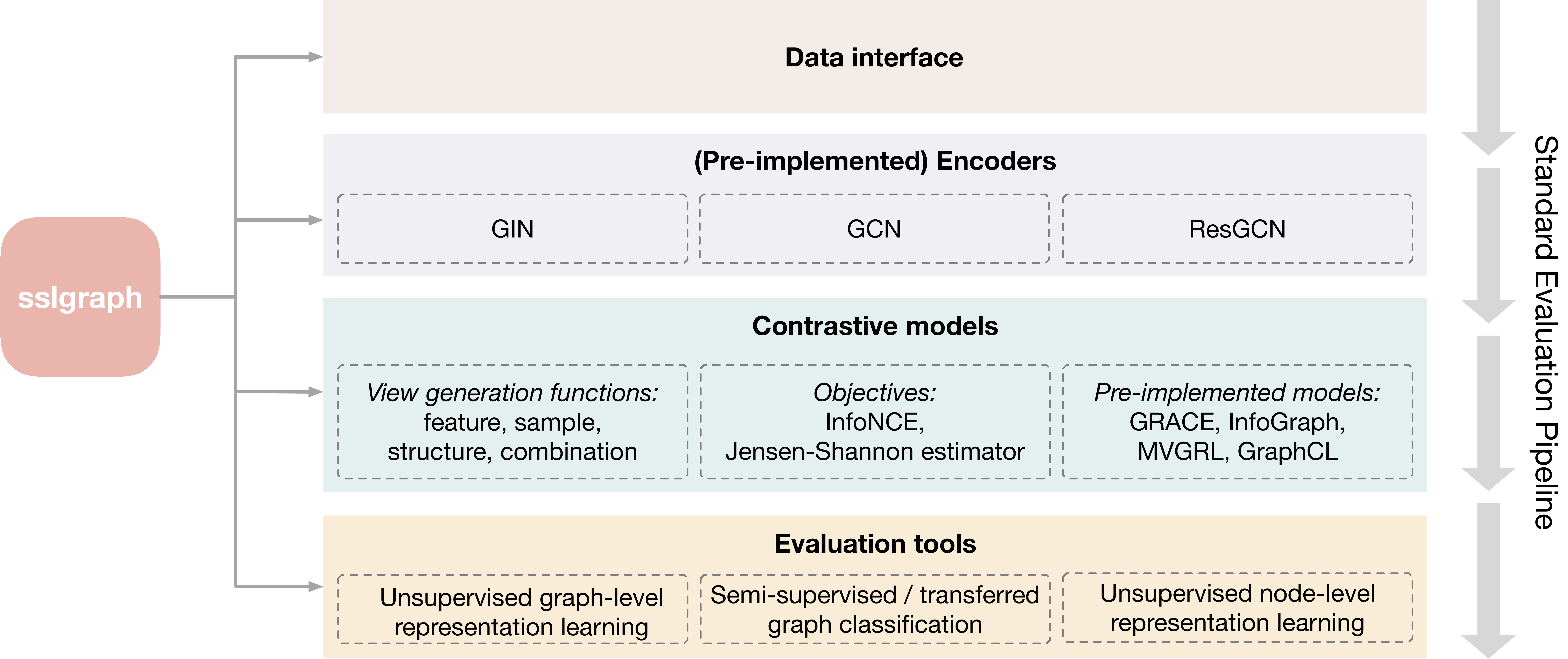}
    \caption{An overview of the developed sslgraph library within DIG: Dive into Graphs. The library provides a standardized evaluation framework consisting of customizable framework and pre-implemented models for contrastive methods, data interface and evaluation tools for both contrastive and predictive methods.}
    \label{fig:sslgraph}
\end{figure*}

\section{Efficiency and Downstream Accuracy of DIG Implementations}

We compare the efficiency and downstream accuracy between individual original implementations and DIG implementations for SSL methods in Supplementary Table~3. All results are obtained from running corresponding implementations under the unsupervised setting on the same environment and device with a single NVIDIA V100 GPU. We use the standard dataset split for training and test~\cite{Fey2019Geometric} for all datasets. Note that the left column of downstream accuracy are reproduced results using the official code provided by the authors and may differ from the original results reported in their paper due to environment difference, randomness, training/test data split, or any unreleased tricks. For example, the original GRACE code uses different datasets split with individual prefixed random seeds for each dataset. To assure fair comparisons, we perform evaluation of the original GRACE under the standard datasets split. Regarding the efficiency, the DIG-sslgraph implementations consume significantly less training time for GraphCL, MVGRL, and InfoGraph, due to more efficient view generation computations. Besides, DIG consumes the same level of GPU memory as original implementations for all four methods. Although DIG-sslgraph does not focus on efficiency optimization, we will keep improving the efficiency in future released versions.

\begin{table*}[ht]\label{tab:dig}
% \normalsize
\centering
\revision{\caption{Comparisons on efficiency and downstream accuracy between individual original implementations and DIG implementations for SSL methods. Efficiencies are compared in terms of in terms of training time and GPU memory. Four pre-implemented methods, GraphCL, MBGRL, InfoGraph, and GRACE, are compared.}
\addtolength{\tabcolsep}{4pt}
\begin{tabular}{ll|cc|cc|cc}
\toprule
Method                     & Dataset  & \multicolumn{2}{c}{\begin{tabular}[c]{@{}c@{}}Time cost per \\ training epoch\end{tabular}} & \multicolumn{2}{c}{\begin{tabular}[c]{@{}c@{}}GPU memory \\ consumption\end{tabular}} & \multicolumn{2}{c}{Downstream accuracy}                \\ \cline{3-8}
                           &          & Original    & DIG       & Original      & DIG       & Original      & DIG\\ \hline
\multirow{3}{*}{GraphCL}   & NCI1     & 68.6915s    & 7.279s    & 1459MB        & 1499MB    & 0.7954 $\pm$ 0.0141 & 0.7961 $\pm$ 0.0143\\
                           & PROTEINS & 5.223s      & 3.996s    & 1463MB        & 1607MB    & 0.7547 $\pm$ 0.0350 & 0.7637 $\pm$ 0.0290\\
                           & MUTAG    & 0.278s      & 0.446s    & 1431MB        & 1475MB    & 0.8991 $\pm$ 0.0495 & 0.9096 $\pm$ 0.0669\\ \hline
\multirow{3}{*}{MVGRL}     & MUTAG    & 9.417s      & 1.602s    & 1917MB        & 2091MB    & 0.8778 $\pm$ 0.0665   & 0.8877 $\pm$ 0.0646\\
                           & PTC-MR   & 8.213s      & 2.986s    & 2827MB        & 2493MB    & 0.5755 $\pm$ 0.0670   & 0.5903 $\pm$ 0.0859\\
                           & IMDB-B   & 27.342s     & 8.557s    & 4459MB        & 4783MB    & 0.7450 $\pm$ 0.0377   & 0.7370 $\pm$ 0.0377\\ \hline
\multirow{3}{*}{InfoGraph} & MUTAG    & 0.585s      & 0.551s    & 1459MB        & 1459MB    & 0.8939 $\pm$ 0.0885   & 0.9041 $\pm$ 0.0927 \\
                           & PTC-MR   & 0.808s      & 0.924s    & 1471MB        & 1445MB    & 0.6249 $\pm$ 0.0966   & 0.6196 $\pm$ 0.0422 \\
                           & IMDB-B   & 9.356s      & 3.013s    & 1485MB        & 1465MB    & 0.7370 $\pm$ 0.0276   & 0.7400 $\pm$ 0.0313\\ \hline
\multirow{3}{*}{GRACE}     & CORA     & 0.020s      & 0.064s    & 1701MB        & 1773MB    & 0.7869 $\pm$ 0.0013   & 0.7877 $\pm$ 0.0096 \\
                           & CiteSeer & 0.030s      & 0.083s    & 2001MB        & 2145MB    & 0.6858 $\pm$ 0.0004   & 0.6842 $\pm$ 0.0061 \\
                           & PubMed   & 0.233s      & 0.345s    & 12703MB       & 13963MB   & 0.8217 $\pm$ 0.0005   & 0.8188 $\pm$ 0.0046\\
\bottomrule
\end{tabular}}
\end{table*}

\end{document}